\documentclass[10pt,twocolumn,letterpaper]{article}

\usepackage{iccv}
\usepackage{times}
\usepackage{epsfig}
\usepackage{graphicx}
\usepackage{amsmath}
\usepackage{amssymb}
\usepackage{comment}
\usepackage{soul}

\usepackage{multirow}
\usepackage{array}
\usepackage{interval}
\newcolumntype{M}[1]{>{\centering\arraybackslash}m{#1}}

\DeclareMathOperator*{\argmin}{arg\,min}

\usepackage{umoline}

\graphicspath{{pics/}}

\usepackage[breaklinks=true,bookmarks=false]{hyperref}

\iccvfinalcopy 

\def\iccvPaperID{5651} 

\ificcvfinal\fi

\begin{document}

\title{Defending Against Universal Perturbations With Shared Adversarial Training}

\author{
	Chaithanya Kumar Mummadi\\
	University of Freiburg\\
	Bosch Center for Artificial Intelligence, Germany\\
	{\tt\small ChaithanyaKumar.Mummadi@de.bosch.com}
	\and
	Thomas Brox\\
	University of Freiburg\\
	{\tt\small brox@cs.uni-freiburg.de}
	\and
	Jan Hendrik Metzen\\
	Bosch Center for Artificial Intelligence, Germany\\
	{\tt\small janhendrik.metzen@de.bosch.com}
}

\maketitle
\ificcvfinal\thispagestyle{empty}\fi

\setcounter{footnote}{1}

\begin{abstract}
	Classifiers such as deep neural networks have been shown to be vulnerable against adversarial perturbations on problems with high-dimensional input space. While adversarial training improves the robustness of image classifiers against such adversarial perturbations, it leaves them sensitive to perturbations on a non-negligible fraction of the inputs. In this work, we show that adversarial training is more effective in preventing universal perturbations, where the same perturbation needs to fool a classifier on many inputs. Moreover, we investigate the trade-off between robustness against universal perturbations and performance on unperturbed data and propose an extension of adversarial training that handles this trade-off more gracefully. We present results for image classification and semantic segmentation to showcase that universal perturbations that fool a model hardened with adversarial training become clearly perceptible and show patterns of the target scene.
\end{abstract}

\section{Introduction}
While deep learning is relatively robust to random noise \cite{fawzi_robustness_2016}, it can be easily fooled by 
\emph{adversarial perturbations} \cite{szegedy_intriguing_2013}. These perturbations are generated by adversarial attacks \cite{goodfellow_explaining_2015,moosavi-dezfooli_deepfool:_2016,carlini_towards_2016} that generate perturbed versions of the
input which are misclassified by a classifier and 
remain quasi-imperceptible for humans. There have been different approaches for explaining properties of adversarial examples and provide rationale for their existence in the first place \cite{goodfellow_explaining_2015,tanay_boundary_2016,fawzi_geometric_nodate,fawzi_classification_2017}. Moreover, these perturbations have been shown to be relatively robust against various kinds of image transformations and are even successful when 
placed as artifacts in the physical world \cite{kurakin_adversarial_2016,sharif_accessorize_2016,evtimov_robust_2017,athalye_synthesizing_2017}. Thus, adversarial perturbations might pose a safety and security risk for autonomous systems and also reduce trust on the models that are in principle vulnerable to these perturbations.
\addtocounter{footnote}{-1}
\begin{figure}[tb]
	\begin{center}
		\text{\scriptsize \hspace{0.7cm}Clean image\hspace{0.6cm} Adv.image undefended model\hspace{0.1cm}Adv.image defended model}
		\includegraphics[width=\linewidth, height=0.22\linewidth]{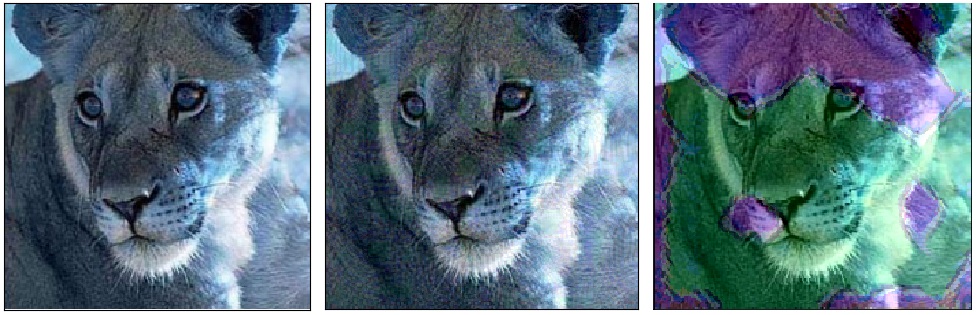}
		\includegraphics[width=\linewidth, height=0.2\linewidth]{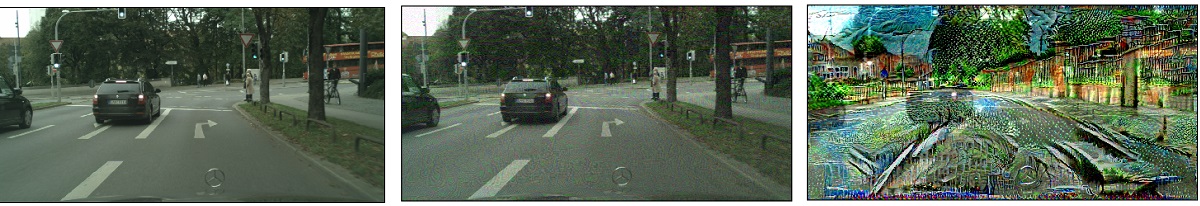}
		
	\end{center}
	\caption{\emph{Effectiveness of shared adversarial training against universal perturbations:} the top row shows an ImageNet example and the bottom row an example from Cityscapes. Adversarial images perturbed by universal perturbations generated for both the undefended models and models defended by our proposed method \emph{shared adversarial training} are shown. The classification accuracy of the defended models deteriorates no more than 5\%  but robustness to universal adversarial attacks increases by 3x and 5x on image classification and semantic segmentation, respectively. Moreover, universal perturbations become clearly perceptible.}
	\label{figure:teaser}
\end{figure}

Several methods have been proposed for increasing the \emph{robustness} of deep networks against adversarial examples, such as adversarial training \cite{goodfellow_explaining_2015,kurakin_adversarial_2017}, virtual adversarial training \cite{miyato_virtual_2017}, ensemble adversarial training \cite{tramer_ensemble_2017}, defensive distillation \cite{papernot_distillation_2016,papernot_extending_2017}, stability training \cite{zheng_improving_2016}, robust optimization \cite{madry_towards_2017}, Parseval networks \cite{cisse_parseval_2017} and alternatively detecting and rejecting them as malicious \cite{metzen_detecting_2017}. While some of these approaches improve robustness against adversarial examples
to some extent, the classifier remains vulnerable against adversarial perturbations on a non-negligible fraction of the inputs for all defenses \cite{athalye_obfuscated_2018,uesato_adversarial_2018}. 

Most work has focused on increasing robustness in image classification tasks, where the adversary can choose a data-dependent perturbation for each input. This setting is very much in favor of the adversary since the adversary can craft a high-dimensional perturbation ``just'' to fool a model on a single input. 
In this work, we argue that limited success in increasing the robustness under these conditions does not necessarily imply that robustness can not be achieved in other settings. Specifically, we focus on robustness against input-agnostic perturbations, namely \emph{universal perturbations} \cite{moosavi-dezfooli_universal_2017}, where the same perturbation needs to fool a classifier on many inputs. Moreover, we investigate robustness against such perturbations in dense prediction tasks such as \emph{semantic image segmentation}, where a perturbation needs to fool a model on many decisions, e.g., the pixel-wise classifications. Data-dependent adversarial attacks need to know their input in advance and require online computation to generate perturbations for every incoming input whereas universal attacks work on unseen inputs.

Prior work has shown that standard models are vulnerable to both universal perturbations, which mislead a classifier on the majority of the inputs \cite{moosavi-dezfooli_universal_2017,mopuri_generalizable_2018}, and to adversarial perturbations on semantic segmentation tasks \cite{fischer_adversarial_2017,xie_adversarial_2017,cisse_houdini:_2017}.

The study of robustness against universal perturbations is important since they pose a realistic threat-model for certain physical-world attacks: for instance, Li et al. \cite{li2019adversarial} show that an adversary could mount a semi-transparent adversarial sticker on a physical camera which effectively adds a universal perturbation to each unseen camera image.
It was demonstrated by Metzen et al. \cite{metzen_universal_2017} that such universal perturbations can hide nearby pedestrians in semantic segmentation which may allow deceiving an emergency braking system and would also pose a threat in surveillance scenarios.
However, these and similar results have been achieved for undefended models.
In this work, we focus on the case where models have been ``hardened" by a defense mechanism, particularly adversarial training. While this technique can considerably increase robustness, there is an implicit trade-off between robustness against perturbations and high performance on unperturbed inputs. We show that explicitly tailoring adversarial training for universal perturbations allows handling this trade-off more gracefully.

Our main contributions are as follows: (1) We propose \emph{shared adversarial training}, an extension of adversarial training that handles the inherent trade-off between accuracy on clean examples and robustness against 
universal perturbations more gracefully.
(2) We evaluate our method on CIFAR10, a subset of ImageNet (with 200 classes), and Cityscapes to demonstrate that universal perturbations for the defended models become clearly perceptible as shown in Figure \ref{figure:teaser}.
(3) We are the first to scale defenses based on adversarial training to semantic segmentation. (4) We demonstrate empirically on CIFAR10 that the proposed technique outperforms other defense mechanisms \cite{moosavi-dezfooli_analysis_2017,perolat_playing_2018} in terms of robustness against universal perturbations.

\section{Related Work}
In this section, we review related work on the study of universal perturbations and adversarial perturbations for semantic image segmentation.

\subsection{Universal Perturbations} \label{section:rw_universal_perturbations}

Different methods for generating universal perturbations exist: Moosavi-Dezfooli et al.\ \cite{moosavi-dezfooli_universal_2017} uses an extension of the DeepFool adversary \cite{moosavi-dezfooli_deepfool:_2016} to generate perturbations that fool a classifier on a maximum number of inputs from a training set. Metzen et al.\ \cite{metzen_universal_2017} proposed a similar extension of the basic iterative adversary \cite{kurakin_adversarial_2017} for generating universal perturbations for semantic image segmentation. In contrast to former works, Mopuri et al.\ \cite{mopuri_fast_2017} proposed Fast Feature Fool, 
a data-independent approach for generating universal perturbations. In follow-up work \cite{mopuri_generalizable_2018}, they show similar fooling rates of data-independent approaches as have been achieved by Moosavi-Dezfooli et al.\ \cite{moosavi-dezfooli_universal_2017}. Khrulkov and Oseledets \cite{khrulkov_art_2017} show a connection between universal perturbations and singular vectors. In another line of work, Hayes and Danezis \cite{hayes_learning_2017}, Mopuri et al. \cite{reddy2018nag}, and Poursaeed et al. \cite{poursaeed2018generative} proposed generative models that can be trained to generate a diverse set of (universal) perturbations.

An analysis of universal perturbation and their properties is provided by Moosavi-Dezfooli et al. \cite{moosavi-dezfooli_analysis_2017}. They connect the robustness to universal perturbations with the geometry of the decision boundary and prove the existence of small universal perturbation provided the decision boundary is systematically positively curved. Jetley et al.\,\cite{jetley_friends_2018} build upon this work and provide evidence that directions in which a classifier is vulnerable to universal perturbations coincide with directions important for correct prediction on unperturbed data. They follow that predictive power and adversarial vulnerability are closely intertwined.

Prior procedures on robustness against universal perturbations define a distribution over (approximately optimal) such perturbations for a model (either by precomputing and random sampling \cite{moosavi-dezfooli_universal_2017}, by learning a generative model \cite{hayes_learning_2017}, or by collecting an increasing set of universal perturbations for model checkpoints during training \cite{perolat_playing_2018}), fine-tune model parameters to become robust against this distribution of 
perturbations, and (optionally) iterate. These procedures increase robustness against universal perturbations slightly, however, not to a satisfying level. This is probably caused by the model overfitting to the fixed distribution of universal perturbations that do not change during the optimization process. However, re-computing universal perturbations in every mini-batch anew is prohibitively expensive. In this work, we propose a method that can be performed efficiently by computing shared perturbations on each mini-batch and using them in adversarial training, i.e., the shared perturbations are computed on-the-fly rather than precomputed as in prior work \cite{moosavi-dezfooli_universal_2017,perolat_playing_2018}. Concurrent to our work, Shafahi et al.\cite{shafahi2018universal} recently proposed ``universal adversarial training'' where updates of the neural network's parameters and the universal perturbation happen concurrently. This reduces the overhead of determining a universal perturbation anew for every mini-batch; however, it is unclear if such an incrementally updated universal perturbation can track the changes of the network's weights sufficiently.

Alternative defense approaches add additional components to the model: Ruan and Dai \cite{ruan_twinnet:_2018} proposed to identify and reject universal perturbations  by adding shadow classifiers, while Akhtar et al.\ \cite{akhtar_defense_2017} proposed to prepend a subnetwork in front of the model that is used to compensate for the added universal perturbation by detecting and rectifying the perturbation. Both methods have the disadvantage that the model becomes large and thus inference more costly. More severely, it is assumed that the adversary is not aware of the defense mechanism and it is unclear if a more powerful adversary could not fool the defense mechanism.

\subsection{Adversarial Perturbations for Semantic Image Segmentation}

Methods for generating adversarial perturbations have been extended to structured and dense prediction tasks like semantic segmentation and object detection
\cite{fischer_adversarial_2017,xie_adversarial_2017,cisse_houdini:_2017}. Metzen et al.\ \cite{metzen_universal_2017} even showed the existence of universal perturbations which result in an arbitrary target segmentation of the scene which has nothing in common with the scene a human perceives. A comparison of the robustness of different network architectures has been conducted by Arnab et al.\ \cite{arnab_robustness_2017}: they found that residual connections and multiscale processing actually increase robustness of an architecture, while mean-field inference for Dense Conditional Random Fields only masks gradient but does not increase robustness itself. In contrast to their work, we focus on modifying the training procedure 
for increasing robustness. Both approaches could be combined in the future.

\section{Preliminaries}
In this section, we introduce basic terms and notations relevant for this work. We aim to defend against an adversary under  \emph{white-box} attack settings. Please refer to Section \ref{section:appendix_threat_model} in the supplementary material for details on capabilities of the adversary and the threat model.

\subsection{Risks}
Let $L$ be a loss function (categorical crossentropy throughout this work), $\mathcal{D}$ be a data distribution, and $\theta$ be the parameters of a parametric model $f_\theta$.
Here, we define the \emph{risk} $\rho(\theta)$ as the expected loss of the model $f_\theta$ for a data distribution.
The following risks are relevant for this work (we extend the definitions of Uesato et al. \cite{uesato_adversarial_2018}):
\begin{enumerate}
	\item Expected Risk: $\rho_{exp}(\theta) = \mathbb{E}_{(x, y) \sim \mathcal{D}} \, L(\theta, x, y)$
	\item Adversarial Risk:\\ $\rho_{adv}(\theta, \mathcal{S}) = \mathbb{E}_{(x, y) \sim \mathcal{D}} \left[\sup\limits_{\xi(x) \in \mathcal{S}} L(\theta, x + \xi(x), y) \right]$
	\item Universal Adversarial Risk:\\ $\rho_{uni}(\theta, \mathcal{S}) = \sup\limits_{\xi \in \mathcal{S}} \mathbb{E}_{(x, y) \sim \mathcal{D}} \left[L(\theta, x + \xi, y) \right]$
\end{enumerate}

Here, $\xi(x)$ denotes an adversarial perturbation, $\xi$ a universal perturbation, and $x + \xi(x)$ an adversarial example. The set $\mathcal{S}$ defines the space from which perturbations may be chosen. 
We would like to note that adversarial and universal risk are not equivalent since in the former case, $\xi(x)$ depends on the specific $x$ sampled from $\mathcal{D}$, while the latter, $\xi$ needs to generalize over the entire data distribution $\mathcal{D}$.

\subsection{Adversaries} \label{sec:adversaries}
Since the worst-case perturbation $\xi(x)$ cannot be computed efficiently in typical settings, one needs to resort to an \emph{adversary} which aims at finding a strong perturbation $\xi(x)$. Note that this corresponds to searching for a tight lower bound of $\rho_{adv}$. We define an adversary as a function $f_{adv}: \mathcal{D} \times \Theta \mapsto \mathcal{S}$, which maps a data point and model parameters $\theta$ onto a perturbation $\xi(x)$ that maximizes a loss $L_{adv} (\theta, x + \xi(x), y)$\footnote{We note that one may choose $L_{adv} = L$ or one may also choose, e.g., $L$ to be the 0-1 loss and $L_{adv}$ be a differentiable surrogate loss.}. 
While different options for the adversaries $f_{adv}$ exist \cite{goodfellow_explaining_2015,moosavi-dezfooli_deepfool:_2016,carlini_towards_2016,moosavi-dezfooli_universal_2017,mopuri_generalizable_2018}, we focus on projected gradient descent (PGD) \cite{madry_towards_2017,kurakin_adversarial_2016}, as it provides in our experience a good trade-off between being computationally efficient and powerful. PGD initializes $\xi^{(0)}$ uniformly at random in $\mathcal{S}$ (or subset of $\mathcal{S}$) and performs $K$ iterations of the following update: 

$\xi^{(k+1)} = \Pi_{\mathcal{S}} \left[\xi^{(k)} + \alpha_k\cdot \text{sgn}(\nabla_x L_{adv}(\theta, x + \xi^{(k)} , y)\right],$
where $\Pi_{\mathcal{S}}$ denotes a projection on the space $\mathcal{S}$ and $\alpha_k$ denotes a step-size. Similarly, a targeted attack where the model shall output the target class $y_{t}$ can be obtained by setting $ \alpha_k$ to $- \alpha_k$ and $y$ to $y_{t}$.

Similar to a standard adversary, we define a \emph{universal adversary} denoted by $f_{uni}$ as function mapping model parameters $\theta$ onto perturbation $\xi$ with the objective of maximizing $\mathbb{E}_{(x, y) \sim \mathcal{D}} \left[L_{adv}(\theta, x + \xi, y) \right]$. One can modify PGD into a universal adversary by using the loss $L_{uni}(\theta, \{x_i, y_i\}_{i=1}^m, \xi) = \frac{1}{m} \sum_{i=1}^m L_{adv}(\theta, x_i + \xi, y_i)$. If the number of data points $m$ is large (which is typically required for finding universal perturbations that generalize well to unseen data), one can employ stochastic PGD, where in every iteration $k$, a set of $\tilde{m}_k$ data points is sampled and $L_{uni}$ is only evaluated on this subset of data points.

\section{Shared Adversarial Training} \label{Section:Methods}
We connect the above risks to show that adversarial training optimizes a loose upper bound on the universal risk and motivate \emph{shared adversarial training}, an extension of adversarial training that aims at maximizing robustness against universal perturbations. We show that this method minimizes an upper bound on the universal risk which is tighter than the one used in adversarial training.

\begin{figure*}[tb]
	\begin{center}
		\includegraphics[width=\linewidth, height=0.23\textwidth]{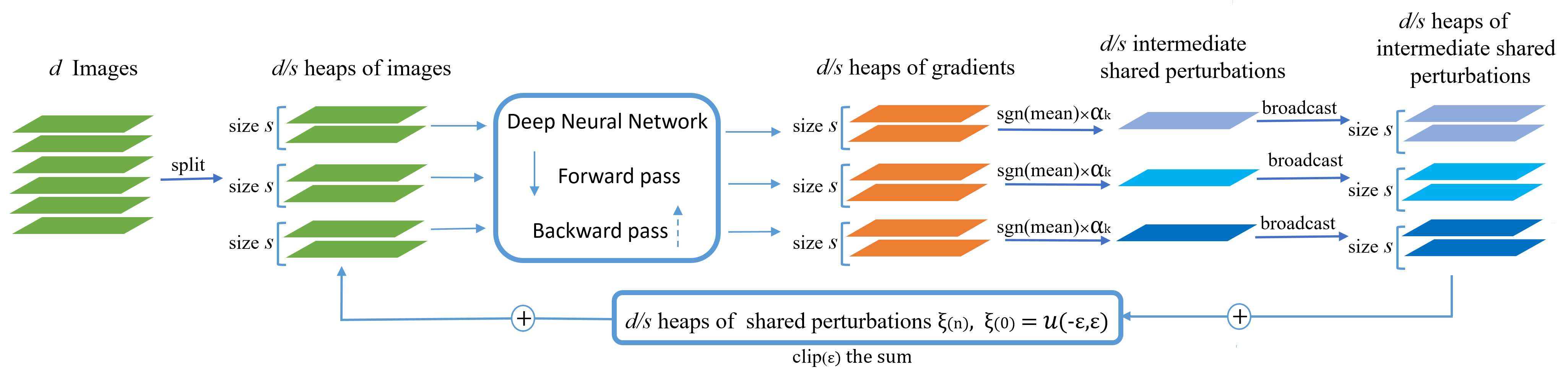}
	\end{center}
	\caption{A pictorial representation of \emph{shared adversarial training}. We split the mini-batch of $d$ images into $d/s$ heaps each with sharedness $s$ and obtain the gradients of the loss with respect to the inputs. Here, the sharedness $s$ corresponds to the number of inputs that are used for the generation of a shared perturbation. The gradients in each heap of size $s$ are then processed and multiplied with step-size $\alpha_k$ to create a shared perturbation that is further broadcasted to size of the heap. The generated shared perturbations are aggregated and clipped after every iteration in order to confine the perturbations within a predefined magnitude $\varepsilon$. These perturbations are added to the images and this process is repeated iteratively. The adversarial inputs generated from the shared perturbations are used for adversarial training.}
	\label{figure:key_concept}
\end{figure*}

\subsection{Relationship between Risks}

We show the following inequalities for the risks:
$\rho_{exp}(\theta) \leq \rho_{uni}(\theta, \mathcal{S}) \leq \rho_{adv}(\theta, \mathcal{S}) \; \forall \theta \; \forall \mathcal{S} \supset \{\mathbf{0}\}$.
To see the validity of these inequalities, we set $ \mathcal{S} = \{\mathbf{0}\}$ to obtain $\rho_{uni}(\theta, \mathcal{S}) = \rho_{exp}(\theta)$ (and $\mathcal{S} \supset \{\mathbf{0}\}$ can only increase $\rho_{uni}(\theta, \mathcal{S})$). 
For the second inequality, assume $\rho_{adv} < \rho_{uni}$. Let $\xi$ be one of the multiple universal perturbations that maximize $\rho_{uni}$. Since $\xi$ is an element of $\mathcal{S}$, we could certainly set $\xi(x) = \xi$ $\forall x$ in the definition of the adversarial risk.
This would result in $\rho_{adv} = \rho_{uni}$.
This completes the proof by contradiction, and thus $\rho_{adv}$ can only be larger than or equal to $\rho_{uni}$ in general.

The objective of \emph{adversarial training} is defined as minimizing the loss function
$\sigma \cdot \rho_{adv}(\theta, \mathcal{S}) + (1 - \sigma) \cdot \rho_{exp}(\theta)$, where $\sigma$ controls the trade-off between robustness and performance on unperturbed inputs. 
We note that if one is interested in minimizing the universal adversarial risk $\rho_{uni}$, then using $\rho_{adv}$ in adversarial training with $\sigma=1$ corresponds to minimizing an upper bound of $\rho_{uni}$ because $\rho_{uni}(\theta, \mathcal{S}) \leq \rho_{adv}(\theta, \mathcal{S})$, provided that the adversaries find perturbations that are sufficiently close to the optimal perturbations. On the other hand, standard empirical risk minimization ERM ($\sigma=0$), which minimizes the empirical estimate of $\rho_{exp}$, corresponds to minimizing a lower bound. As shown in previous work \cite{goodfellow_explaining_2015,moosavi-dezfooli_deepfool:_2016,carlini_towards_2016}, this does confer only little robustness against (universal) perturbations.  For $0 < \sigma < 1$, adversarial training corresponds to minimizing a convex combination
of the upper bound $\rho_{adv}$ and the lower bound $\rho_{exp}$ but does not directly optimize on $\rho_{uni}$. As we show in Section \ref{Section:Experiments}, this standard version of adversarial training already provides strong robustness against universal perturbations at the cost of reducing performance on unperturbed data considerably. 

\subsection{Method}
Directly employing $\rho_{uni}$ in adversarial training is infeasible since evaluating $\rho_{uni}(\theta, \mathcal{S})$ with an adversary $f_{uni}$ in every mini-batch is prohibitively expensive (because it requires large $m$). Hence, it would be desirable to use an upper bound of $\rho_{uni}$ in adversarial training that is tighter than $\rho_{adv}$ but cheaper to approximate than $\rho_{uni}$.

For this, we propose to use a so-called \emph{heap adversary}, which we define as a function $f_{heap}: \mathcal{D}^m \times \Theta \mapsto \mathcal{S}$ that maps a set of $m$ data points and model parameters $\theta$ onto a perturbation $\xi$. We use $L_{uni}(\theta, \{x_i, y_i\}_{i=1}^m, \xi) = \frac{1}{m} \sum_{i=1}^m L_{adv}(\theta, x_i + \xi, y_i)$ as loss function for the heap adversary. However, in contrast to a universal adversary, we do not require a heap adversary to find perturbations that generalize to unseen data. This allows choosing $m$ relatively small.

More specifically, we split a mini-batch consisting of $d$ data points into $d / s$ heaps (subsets of the mini-batch) of size $s$ (we denote $s$ as \emph{sharedness}). Rather than using the adversary $f_{adv}$ for computing a perturbation on each of the $d$ data points separately, we employ a heap adversary $f_{heap}$ for computing $d / s$ \emph{shared} perturbations on the heaps with $m = s$. Thereupon, these perturbations are broadcasted to all $d$ data points by repeating each of the shared perturbations $s$ times for all elements of the heap. Employing this heap adversary implies a risk $\rho_{heap}^{(s)}$. We propose to use $\rho_{heap}^{(s)}$ in adversarial training when aiming at defending against universal perturbations and denote the resulting procedure as \emph{shared adversarial training}. This entire process is illustrated in Figure \ref{figure:key_concept}. We can obtain the following relationship for $s = 2^{i}$ (please refer to Section \ref{section:sharedness_relationship} for more details):\\
$$\rho_{adv} = \rho_{heap}^{(1)}
\geq \rho_{heap}^{(2)}
\geq \rho_{heap}^{(4)}
\geq \dots
\geq \rho_{heap}^{(d)}
\geq \rho_{uni}(\sigma, \mathcal{S})$$

Note that while all $\rho_{heap}^{(s)}$ are upper bounds on the universal risk $\rho_{uni}$, this does not imply that shared perturbations are strong universal perturbations. In contrast, the smaller $s$, the more ``overfit'' are the shared perturbations to the respective heap. However, $\rho_{heap}^{(s)}$ with $s \gg 1$ is typically a much tighter upper bound on $\rho_{uni}$ than $\rho_{adv}$ and can be approximated as efficiently as $\rho_{adv}$: for this, PGD is converted into a heap adversary by replacing $L_{adv}$ with $L_{uni}$. By appropriately reshaping and broadcasting perturbations, we can compute $d / s$ shared perturbations on the respective heaps of the mini-batch jointly by PGD with essentially the same cost as computing $d$ adversarial perturbations with PGD.

\subsection{Adversarial Loss Function}
We recall that $L_{uni}(\theta, \{x_i, y_i\}_{i=1}^m, \xi) = \frac{1}{m}\sum_{i=1}^m L_{adv}(\theta, x_i + \xi, y_i)$. Because of limited capacity of the perturbation ($\xi \in \mathcal{S}$), there is ``competition'' between $m$ data points: the maximizers of $L_{adv}(\theta, x_i + \xi, y_i)$ will typically be different and the data points will ``pull'' $\xi$ into different directions. Hence, using the categorical cross-entropy as a proxy for the 0-1 loss is problematic for untargeted adversaries: since we are maximizing the loss and the categorical cross-entropy has no upper bound, there is a winner-takes-all tendency where the perturbation is chosen such that it leads to highly confident misclassifications on some data points and to correct classification on other data points (this incurs higher cost than misclassifying more data points but with lower confidence).

To prevent this, we employ loss thresholding on the categorical cross-entropy $L$ to enforce an upper bound on $L_{adv}$: $L_{adv}(\theta, x, y) = \min(L(\theta, x, y), \kappa)$.  We used $\kappa = - \log 0.2$, which corresponds to not encouraging the adversary to reduce confidence of the correct class below $0.2$. 
A similar loss thresholding was also proposed by Shafahi et al. \cite{shafahi2018universal} concurrently.
Besides, we also incorporate label smoothing and use the soft targets for the computation of loss in all our experiments.

\section{Robustness Evaluation} \label{Section:Robustness}
In this section, we define the measure of robustness used in the experiments and detail how we approximate it.

\subsection{Definition of Robustness}

For the special case of the 0-1 loss, an $n$-dimensional input $x$, and $\mathcal{S} = \mathcal{S}(\varepsilon) = [-\varepsilon, \varepsilon]^n$, we define the \emph{adversarial robustness} as  the smallest perturbation magnitude $\varepsilon$ that results in an adversarial risk (misclassification rate) of at least $\delta$. More formally:\\
$\varepsilon_{adv}(\delta) = \argmin\limits_{\varepsilon} \rho_{adv}(\theta, \mathcal{S}(\epsilon)) \text{ s.t. } \rho_{adv}(\theta, \mathcal{S}(\epsilon)) > \delta.$\\
In other words, there are perturbations $\xi(x)$ with $\vert\vert \xi(x)\vert\vert_\infty < \varepsilon_{adv}(\delta)$ that result in a misclassification rate of at least $\delta$. Analogously, we can also define the \emph{universal robustness} as\\
$\varepsilon_{uni}(\delta) = \argmin\limits_{\varepsilon} \rho_{uni}(\theta, \mathcal{S}(\epsilon)) \text{ s.t. } \rho_{uni}(\theta, \mathcal{S}(\epsilon)) > \delta.$\\
Here, a perturbation $\xi$ with $\vert\vert \xi\vert\vert_\infty < \varepsilon_{uni}(\delta)$ exists that results in a misclassification rate of at least $\delta$.

\subsection{Quantifying Robustness} 
\begin{figure*}[tb]
	\begin{center}
		\includegraphics[width=.38\textwidth, height=0.28\textwidth]{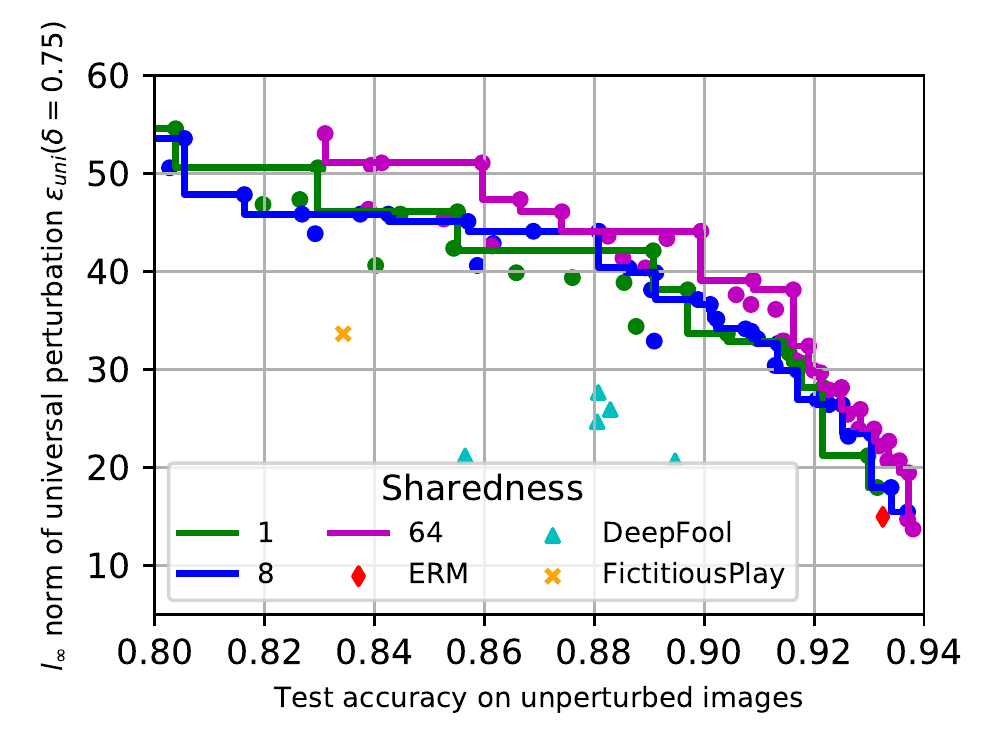}
		\includegraphics[width=.38\textwidth, height=0.28\textwidth]{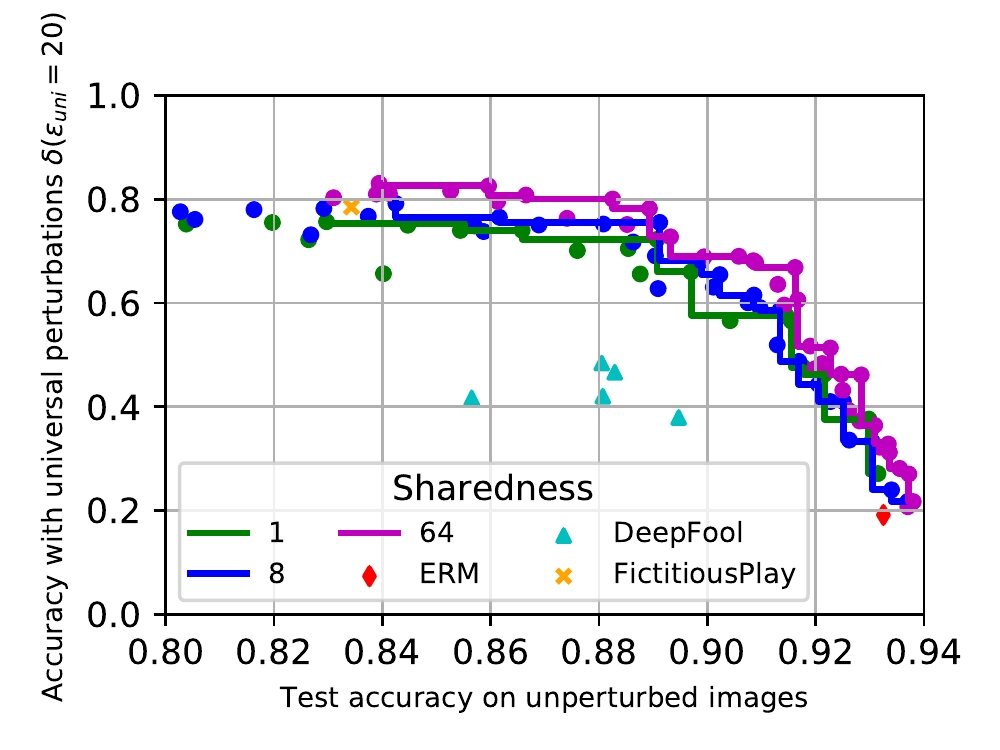}
	\end{center}
	\caption{Pareto front on CIFAR10 for sharedness values $s \in \{1, 8, 64\}$. ERM corresponds to the model pretrained with empirical risk minimization, ``DeepFool UAD" \cite{moosavi-dezfooli_universal_2017} to models trained with the procedure proposed by Moosavi-Dezfooli et al. \cite{moosavi-dezfooli_universal_2017}, and ``FictitiousPlay" to the procedure proposed by Perolat et al. \cite{perolat_playing_2018}. (Left) Robustness with regard to S-PGD universal perturbations.  (Right) Robustness with regard to DeepFool-based universal perturbations \cite{moosavi-dezfooli_universal_2017}. The Pareto front of the proposed defense is clearly above all previous defenses.}
	\label{figure:cifar_pareto}
\end{figure*}
Since the exact evaluation of $\varepsilon_{uni}(\delta)$ is intractable for our settings, we use an upper bound on the actual robustness $\varepsilon_{uni}(\delta)$ instead.
For this, we tuned the PGD adversary as follows to make it more powerful (and thus the upper bound more tight): we performed a binary search of $b$ iterations for perturbation magnitude $\varepsilon$ of $\mathcal{S}(\varepsilon)$, i.e., the bound in the $l_\infty$ norm on the perturbation, on the interval $\varepsilon \in [0, 255]$. In every iteration, we used the step-size annealing schedule $\alpha_k = \frac{\beta \varepsilon \gamma^k}{\sum_{j=0}^{K-1} \gamma^j}$ which guarantees that $\sum_{j=0}^{K-1} \alpha_k = \beta \varepsilon$. If a perturbation with misclassification rate $\delta$ is found in an iteration, the next iteration of binary search continues on the lower half of the interval for $\varepsilon$, otherwise on the upper half. The reported robustness is the smallest perturbation found in entire procedure that achieves a misclassification rate of $\delta$. Note that this procedure was only used for evaluation; for training we used a predefined $\varepsilon$ and constant step-size $\alpha_k$.

\section{Experimental Results} \label{Section:Experiments}
We present experimental results of \emph{shared adversarial training} on robustness against universal perturbations in both image classification and semantic segmentation tasks. We extended the PGD implementation of Cleverhans \cite{papernot2016cleverhans} such that it supports shared adversarial perturbations and loss clipping as discussed in Section \ref{Section:Methods}. For quantifying robustness, we extended Foolbox \cite{rauber2017foolbox} such that universal perturbations (with minimal $l_\infty$ norm) that achieve a misclassification rate of at least $\delta$ can be searched.

\subsection{Experiments on CIFAR10}

We present results on CIFAR10 \cite{krizhevsky_learning:2009} for ResNet20 \cite{DBLP:journals/corr/HeZRS15} with 64-128-256 feature maps per stage. For evaluating robustness, we generate $f_{uni}$ using stochastic PGD on 5000 validation samples with mini-batches of size $\tilde{m}_k = 16$ and evaluated on $512$ test samples. We used $b=10$ binary search iterations, $K=200$ S-PGD iterations, and the step-size schedule values $\gamma=0.975$ and $\beta=4$. We pretrained ResNet20 with standard regularized empirical risk minimization (ERM) and obtained an accuracy of $93.25\%$ on clean data and a robustness
against universal perturbations of $\varepsilon_{uni}(\delta=0.75) = 14.9$. 

In general, we are interested in models that increase the robustness without decreasing the accuracy on clean data considerably. We consider this as a multi-objective problems with two objectives (accuracy and robustness). In order to approximate the Pareto-front of different variants of adversarial and shared adversarial training (sharedness $s \in \{1, 8, 64\}$), we conducted runs for a range of attack parameters: maximum perturbation strength $\varepsilon \in \{2, 4, 6, 8, 10, 14, 18, 22, 26\}$ and $\sigma \in \{0.3, 0.5, 0.7, 0.9\}$ (controlling the trade-off between expected and adversarial risk).
Model fine-tuning was performed with 65 epochs of SGD with batch-size 128, momentum $0.9$, initial learning rate of $0.0025$ and also performed 4 steps of PGD with step-size $\alpha_k = 0.5 \varepsilon$ for each mini-batch. Here, the learning rate was annealed after 50 epochs by a factor of 10. 

Figure \ref{figure:cifar_pareto} (left) shows the resulting Pareto fronts of different sharedness values (entries are provided in Table \ref{tab:cifar10_robustness} in supplementary material). While sharedness $s=1$ (standard adversarial training) and $s=8$ perform similarly, $s=64$ strictly dominates the other two settings. Without any loss on accuracy, a robustness of $\varepsilon_{uni}(\delta=0.75) = 22.7$ can be achieved, and if one accepts an accuracy of $90\%$, a robustness of $\varepsilon_{uni}(\delta=0.75) = 44.1$ is obtainable. This corresponds to nearly three times the robustness of the undefended model while accuracy only drops by less than $3.5\%$. We would also like to note that standard adversarial training is surprisingly effective in defending against universal perturbations and achieves a robustness that is smaller by approximately 5 than $s=64$ at the same level of accuracy on unperturbed data. These findings suggest that increasing sharedness results in increased robustness. We found in preliminary experiments that this effect is strong for small $s$ but has diminishing returns for sharedness beyond $s=64$.

We also evaluated the defenses against universal perturbations proposed by Moosavi-Dezfooli et al. \cite{moosavi-dezfooli_universal_2017} and Perolat et al. \cite{perolat_playing_2018} (please refer to Section \ref{section:appendix_baseline_configuration} in the supplementary material for details). It can be seen in Figure \ref{figure:cifar_pareto} (left) that these defenses are strictly dominated by all variants of (shared) adversarial training. In terms of computation, shared adversarial training required 189s (the same compute time as required by standard adversarial training) while the defense \cite{moosavi-dezfooli_universal_2017} required 3118s, and the defense \cite{perolat_playing_2018} required 3840s per epoch on average. The proposed method thus outperforms the baseline defenses both in terms of computation and with regard to the robustness-accuracy trade-off.

Figure \ref{figure:cifar_pareto} (right) shows the Pareto front of the same models when attacked by the DeepFool-based method for generating universal perturbations \cite{moosavi-dezfooli_universal_2017}. In this case, the robustness is computed for a fixed perturbation magnitude $\varepsilon_{uni}=20$ and the accuracy $\delta$ under this perturbation is reported. The qualitative results are the same as for an S-PGD attack: the Pareto-front of adversarial training (s=1) clearly dominates the results achieved by the defense proposed in \cite{moosavi-dezfooli_universal_2017}. Moreover, shared adversarial training with s=64 dominates standard adversarial training and the defense proposed by Perolat et al. \cite{perolat_playing_2018}. This indicates that the increased robustness by shared adversarial training is not specific to the way the attacker generates universal perturbations. An illustration of the universal perturbations on this dataset is given in Section \ref{section:cifar10_perturbations} in the supplementary material.

\subsection{Experiment on a Subset of ImageNet}
We extend our experiments to a subset of ImageNet \cite{deng2009imagenet}, which has more classes and higher resolution inputs than CIFAR10.
Please refer to Section \ref{section:subset_imagenet} in the supplementary material for details on the selection of this subset.
Similar to CIFAR10, we evaluate the robustness using stochastic PGD but generate perturbations on the training set with mini-batches of size $\tilde{m}_k = 10,000$ and evaluate on the total validation set. We used $b=10$ binary search iterations, $K=20$ S-PGD iterations, and the step-size schedule values $\gamma=0.975$ and $\beta=4$. We pre-trained wide residual network WRN-50-2-bottleneck \cite{zagoruyko2016wide} on this dataset with ERM using SGD for 100 epochs along with initial learning rate 0.1 and reduced it by a factor of 10 after every 30 epochs. We have obtained a top-1 accuracy of 77.57\%  on unperturbed validation data and a robustness
against universal perturbations of $\varepsilon_{uni}(\delta=0.75) = 8.4$.

We approximate the Pareto front of adversarial and shared adversarial training with sharedness $s \in \{1, 32\}$ and different $\varepsilon \in \{2, 4, 6, 8, 10, 14, 18, 22, 26\}$ and $\sigma \in \{0.5, 1.0\}$. We performed 5 steps of PGD with step-size $\alpha_k = 0.4 \varepsilon$. The model was fine-tuned for 30 epochs of SGD with batch-size 128, momentum term $0.9$, weight decay $5e^{-5}$, an initial learning rate of $0.01$ that was reduced by a factor of 10 after 20 epochs and also performed 5 steps of PGD with step-size $\alpha_k = 0.4 \varepsilon$ for each mini-batch.

\begin{figure}[tb]
	\begin{center}
		\includegraphics[width=.38\textwidth, height=.17\textwidth]{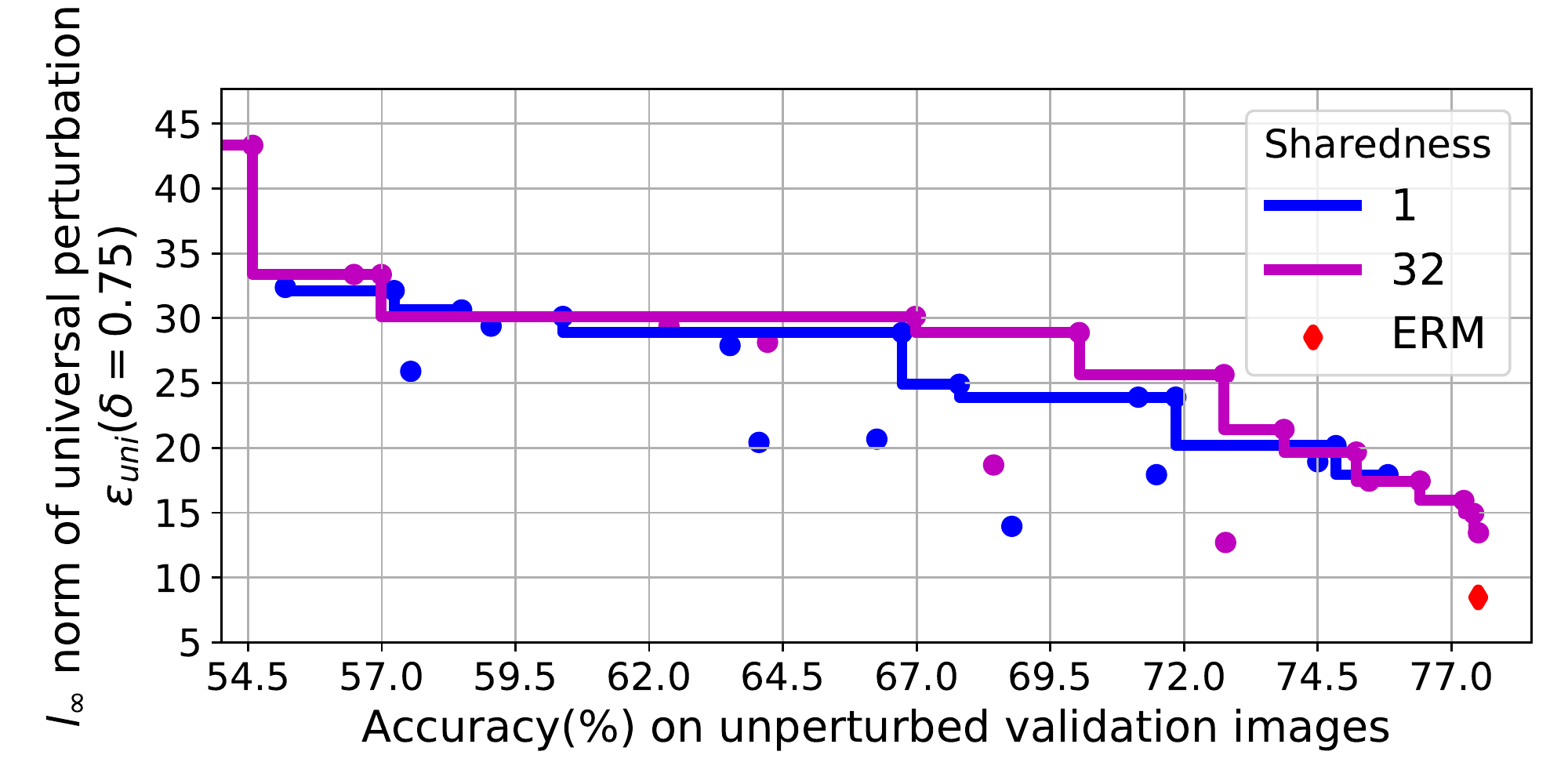}
		
	\end{center}
	\caption{Pareto front on ImageNet for sharedness $s \in \{1, 32\}$. Shared adversarial training has doubled the robustness at the point of accuracy similar to baseline. With a slight loss of accuracy between $5\%$ to $7\%$, the method increases the robustness by a factor of 3 and clearly dominates the standard adversarial training in terms of the robustness/accuracy trade-off.
	}
	\label{figure:tinyimagenet_pareto}
\end{figure}

Figure \ref{figure:tinyimagenet_pareto} compares the Pareto front of shared adversarial training with $s=32$ and standard adversarial training $s=1$ (entries are provided in Table \ref{tab:imagenet_robustness} in the supplementary material). It can be clearly seen that shared adversarial training increases the robustness from $\varepsilon_{uni}(\delta=0.75) = 8.4$ to $15.0$ without any loss of accuracy. Moreover, shared adversarial training also dominates standard adversarial training for a target accuracy between 67\%-74\%, which corresponds to the sweet spot as a small loss in accuracy allows a large increase in robustness. The point with accuracy 72.74\% and robustness $\varepsilon_{uni}(\delta=0.75) = 25.64$ (obtained at $s=32, \varepsilon=10, \sigma=1.0$) can be considered a good trade-off as accuracy drops by only 5\% while robustness increases by a factor of 3, which results in clearly perceptible perturbations as shown in the top row of Figure \ref{figure:teaser} and Section \ref{section:appendix_imagenet_perturbations}.
Moreover, (shared) adversarial training also increases the entropy of the predicted class distribution for successful untargeted perturbations substantially (see Section \ref{section:appendix_imagenet_class_count}).

\subsection{Semantic Image Segmentation} \label{Section:Semseg}
The results from above experiments show that shared adversarial training improves robustness against universal perturbations on image classification tasks where the adversary aims to fool the classifier's single decision on an input. In this section, we investigate our method against adversaries in a dense prediction task (semantic image segmentation), where the adversary aims at fooling the classifier on many decisions. To our knowledge,  this is the first work to scale defenses based on adversarial training to this task. 

We evaluate the proposed method on the Cityscapes dataset \cite{Cordts2016Cityscapes}. For computational reasons, all images and labels were downsampled from a resolution of $2048\times1024$ to $1024\times512$ pixels, where for images a bilinear interpolation and for labels a nearest-neighbor approach was used for downsampling. We pretrained the FCN-8 network architecture \cite{long_fully_2015} on the whole training set of $2975$ images and achieved 49.3\% class-wise intersection-over-union (IoU) on the validation set of $500$ images. Note that this IoU is relatively low because of downsampling the images.

We follow the experimental setup of Metzen et al. \cite{metzen_universal_2017} which performed a targeted attack with a fixed target scene (monchengladbach\_000000\_026602\_gtFine). They demonstrated that the desired target segmentation can be achieved despite the fact that the original scene has nothing in common with the target scene. We use the same target scene and consider this targeted attack successful if the average pixel-wise accuracy between the prediction on the perturbed images and the target segmentation exceeds $\delta=0.95$. For evaluating robustness, we generate $f_{uni}$ using stochastic PGD with mini-batches from the validation set of size $\tilde{m}_k = 5$ and tested on $16$ samples from test set. We used $b=10$ binary search iterations, $K=200$ S-PGD iterations, the step-size schedule values $\gamma=0.99$ and $\beta=2$, and did not employ loss thresholding for targeted attacks. We find a universal perturbation that upper bounds the robustness of the model to $\varepsilon_{uni}(\delta=0.95) \leq 19.92$.

\begin{figure}[tb]
	\begin{center}
		\includegraphics[width=\linewidth]{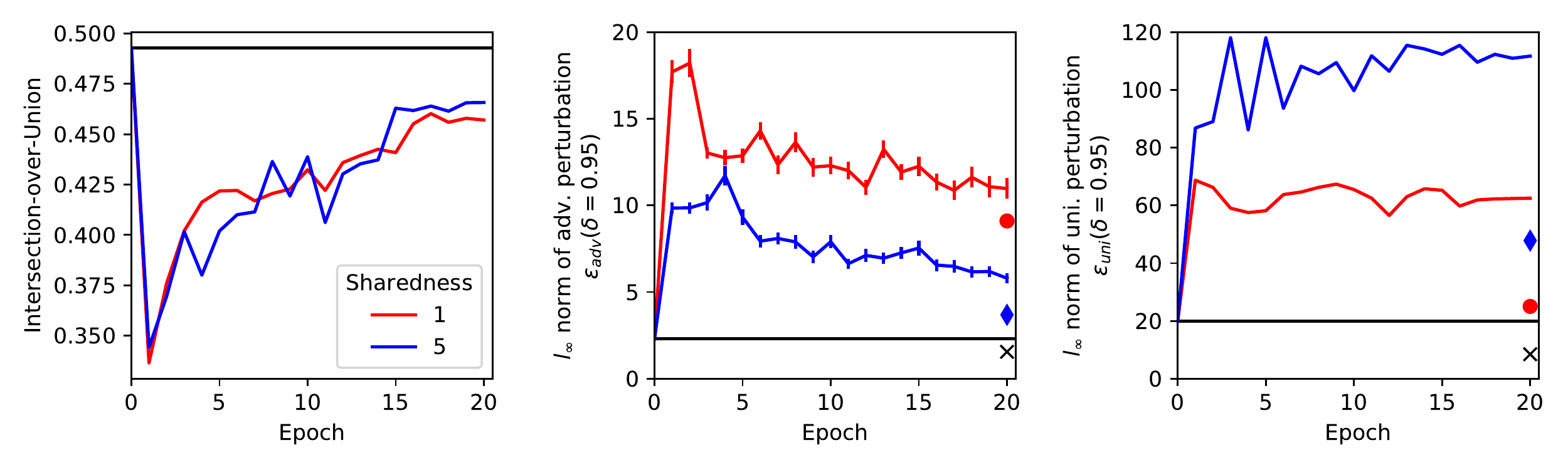}
	\end{center}
	\caption{Learning curves on Cityscapes for adversarial (red, circle) and shared adversarial training (blue, diamond) with regard to performance on unperturbed images (left), and robustness against adversarial perturbations (middle, showing mean and standard error of mean) and universal perturbations (right). Black horizontal lines denote performance of undefended model. Isolated markers correspond to robustness against untargeted attacks. Performance of both standard and shared adversarial training are comparable on unperturbed data, but standard adversarial training dominates in terms of robustness against image-dependent adversarial perturbations, while shared adversarial training dominates in terms of robustness against targeted and untargeted universal perturbations.}
	\label{figure:cityscapes_learning_curve}
\end{figure}

We fine-tuned this model with adversarial and shared adversarial training. Since approximating the entire Pareto front of both methods would be computationally very expensive, we instead selected a target performance on unperturbed data of roughly 45\% IoU (no more than $5\%$ worse than the undefended model).  The following two settings achieved this target performance
(see Figure \ref{figure:cityscapes_learning_curve} left): adversarial training with $\varepsilon=8$ and $\sigma=0.5$ and shared adversarial training for sharedness $s = 5$, $\varepsilon=30$, and $\sigma=0.7$.
The finetuning was performed for 20 epochs using Adam with batch-size 5 and a learning rate of $0.0001$ that was annealed after 15 epochs to $0.00001$.
As heap adversary, we performed 5 steps of untargeted PGD with step-size $\alpha_k = 0.4 \varepsilon$. 

While both methods achieved very similar performance on unperturbed data, they show very different robustness against adversarial and universal perturbations (see Figure \ref{figure:cityscapes_learning_curve}): standard adversarial training largely increases robustness against adversarial perturbations to $\varepsilon_{adv}(\delta=0.95) \leq 11$, an increase by a factor of 4 compared to the undefended model. Shared adversarial training is less effective against adversarial perturbations, its robustness is upper bounded by $\varepsilon_{adv}(\delta=0.95) \leq 5.9$. However, shared adversarial training is more effective against targeted universal perturbations with an upper bound on robustness of $\varepsilon_{uni}(\delta=0.95) \leq 111.7$, while adversarial training reaches $\varepsilon_{uni}(\delta=0.95) \leq 62.5$.

We also evaluated robustness against untargeted attacks: robustness increased from  $\varepsilon_{uni}(\delta=0.95) \leq 8.5$ of the undefended model to
$25$ and $47.8$ for the models trained with standard and shared adversarial training respectively. The universal perturbation for the model trained with shared adversarial training clearly shows patterns of the target scene and dominates the original image, which is also depicted in the bottom row of Figure \ref{figure:teaser}. We refer to Section \ref{section:cityscapes_adv_examples} in the supplementary material for illustrations of targeted and untargeted universal perturbations for different models.

\subsection{Discussion}

\begin{figure}[tb]
	\begin{center}
		\includegraphics[width=\linewidth, height=0.46\linewidth]{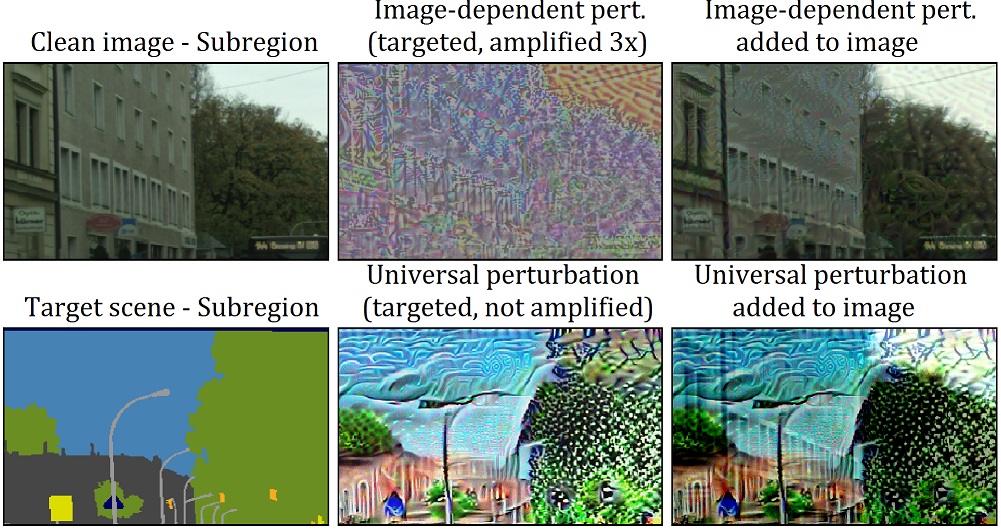}
	\end{center}
	\caption{Illustration of image-dependent and universal perturbations for the same image and target scene (upper and lower left) that are generated on the model hardened with shared adversarial training.
		Image-dependent perturbations weaken patterns of existing structure like edges of the actual scene (upper right) whereas universal perturbations are restricted to adding structure indicative of the target scene (lower right). This qualitative difference between perturbations provides a possible explanation why 
		shared adversarial training demonstrates different levels of robustness on image-dependent and universal perturbations: 
		shared adversarial training improves robustnesss against additive structure 
		but not against the perturbations that weaken the existing structure.}
	\label{figure:cityscapes_zoomin}
\end{figure}

Results shown in Figure \ref{figure:cityscapes_learning_curve} indicate that there may exist a trade-off between robustness against image-dependent adversarial perturbations and universal perturbations. Figure \ref{figure:cityscapes_zoomin} illustrates why these two kinds of robustness are not strictly related: adversarial perturbations fool a classifier by both adding structure from the target scene/class\footnote{For untargeted attacks, the attacks may choose a target scene/class arbitrarily such that fooling the model becomes as simple as possible.} to the image (e.g., 
vegetation on the middle left part of the image) and by destroying properties of the original scene (e.g., edges of the topmost windowsills). The latter is not possible for universal perturbations since the input images are not known in advance. As also shown in the figure, universal perturbations compensate this by adding stronger patterns of the target scene. Shared perturbations will become more similar to universal perturbations with increasing sharedness since a single shared perturbation has fixed capacity and cannot destroy properties of arbitrarily many input images (even if they are known in advance). Accordingly, shared adversarial training will make the model mostly more robust against perturbations which add new structures and not against perturbations which destroy existing structure. Hence, it results in less robustness against image-specific perturbations (seen in Figure \ref{figure:cityscapes_learning_curve} middle). On the other hand, since shared adversarial training focuses on one specific kind of perturbations (those that add structure to the scene), it leads to models that are particularly robust against universal perturbations (shown in Figure \ref{figure:cityscapes_learning_curve} right).

\section{Conclusion}
We have shown that adversarial training is surprisingly effective in defending against universal perturbations. Since adversarial training does not explicitly optimize the trade-off between robustness against universal perturbations and performance on unperturbed data points, it handles this trade-off suboptimally. We have proposed \emph{shared adversarial training}, which performs adversarial training on a tight upper bound of the universal adversarial risk. We have shown that our method allows achieving high robustness against universal perturbations on image classification tasks at smaller loss of accuracy. The proposed method also scales to semantic segmentation on high resolution images, where compared to adversarial training it achieves higher robustness against universal perturbations at the same level of performance on unperturbed images.

\clearpage

{\small
	\bibliographystyle{ieee_fullname}
	\bibliography{references}
}

\newpage
\clearpage

\onecolumn
\begin{center}
	\Large\textbf{Defending Against Universal Perturbations With Shared Adversarial Training}\\
	\large\text{}\\
\end{center}

\renewcommand{\thetable}{A\arabic{table}}
\renewcommand{\thefigure}{A\arabic{figure}}
\setcounter{table}{0}
\setcounter{figure}{0}
\setcounter{section}{0}
\setcounter{page}{1}

\appendix
\section{Supplementary material}
\setcounter{footnote}{0}
\subsection{Threat Model}
\label{section:appendix_threat_model}
Here, we specify the capabilities of the adversary since the proposed defense mechanism aims at providing security under a specific threat model. We assume a \emph{white-box} setting, where the adversary has full information about the model, i.e., it knows network architecture and weights, and can provide arbitrary inputs to the model and observe their corresponding outputs (and loss gradients). Moreover, we assume that the attacker can arbitrarily modify every pixel of the input but aims at keeping the $l_\infty$ norm of this perturbation minimal. In the case of a universal perturbation, we assume that the attacker can choose an arbitrary perturbation (while aiming to keep the $l_\infty$ norm minimal), but crucially does not know the inputs to which this perturbation will be applied. The adversary, however, has access to data points that have been sampled from the same data distribution as the future inputs.

\subsection{Relationship of different sharedness}
\label{section:sharedness_relationship}
Provided that the heap adversary finds a perturbation that is sufficiently close to the optimal perturbation of the heap and that heaps are composed hierarchically\footnote{Heaps are composed hierarchically when a heap of sharedness $2s$ is always the union of two disjoint heaps of sharedness $s$.}, we have the following relationship for $s = 2^{i}$ (we omit the dependence on $\sigma$, $\mathcal{S}$ and $f_{adv}$/$f_{heap}$/$f_{uni}$ for brevity):

$$\tilde{\rho}_{adv} = \tilde{\rho}_{heap}^{(1)}
\geq \tilde{\rho}_{heap}^{(2)}
\geq \tilde{\rho}_{heap}^{(4)}
\geq \dots
\geq \tilde{\rho}_{heap}^{(d)}
\geq \tilde{\rho}_{uni}(\sigma, \mathcal{S})$$

To see $\tilde{\rho}_{heap}^{(s)} \geq \tilde{\rho}_{heap}^{(2s)}$, let $\xi_1, \dots, \xi_{d/(2s)}$ be the shared perturbations on the $d/(2s)$ heaps of $\tilde{\rho}_{heap}^{(2s)}$. Let $\xi_j$ be the shared perturbation for the $j$-th heap. Then, because of the hierarchical construction of the heaps, this heap is composed of two heaps used in 
$\tilde{\rho}_{heap}^{(s)}$. Let $j_1$ and $j_2$ be the corresponding indices of these heaps in $\tilde{\rho}_{heap}^{(s)}$. By setting $\xi_{j_1} = \xi_{j_2} = \xi_j$, we obtain $\tilde{\rho}_{heap}^{(s)} = \tilde{\rho}_{heap}^{(2s)}$.

\subsection{Configuration of Baselines for CIFAR10} \label{section:appendix_baseline_configuration}
For the defense proposed by Moosavi-Dezfooli et al.\ \cite{moosavi-dezfooli_universal_2017}, we generated 10 different universal perturbations using the DeepFool-based method for generating universal-perturbations on 10,000 randomly sampled training images, ran 5 epochs of adversarial training with $\sigma=0.5$, and chose the applied perturbation uniform randomly from the precomputed perturbations. After these 5 epochs, the robustness was evaluated. This procedure was iterated five times, which resulted in 5 accuracy-robustness points in Figure 1.

We run the defense proposed by Perolat et al. \cite{perolat_playing_2018} for 45 epochs (sufficiently long for achieving convergence as evidenced by Figure 4 of Perolat et al. \cite{perolat_playing_2018}). At the beginning of each episode, we generated one universal perturbation using the DeepFool-based method for generating universal-perturbations on the entire training set. We used $\sigma=0.25$, and chose the applied perturbation uniform randomly from all universal perturbations computed so far. We report the accuracy and robustness at the end of these 45 epochs. We note that even though we did not replicate the exact setup of Perolat et al. \cite{perolat_playing_2018}, we achieve a similar accuracy-robustness trade-off in Figure \ref{figure:cifar_pareto} (right) as the one given in Figure 4 of Perolat et al. \cite{perolat_playing_2018}.

\subsection{Illustration of Universal Perturbations on CIFAR10} \label{section:cifar10_perturbations}

\begin{figure*}[b]
	\begin{center}
		\includegraphics[width=.9\linewidth]{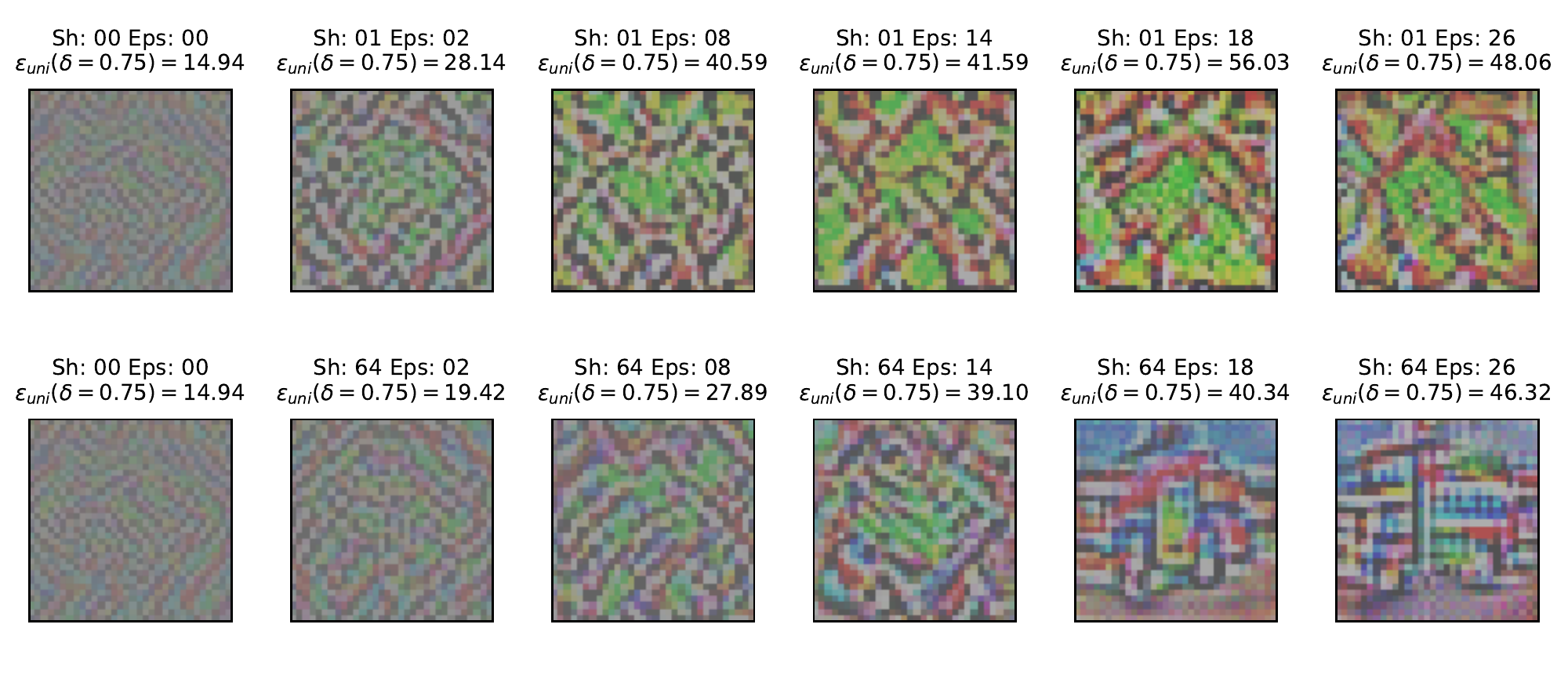}
		\includegraphics[width=.9\linewidth]{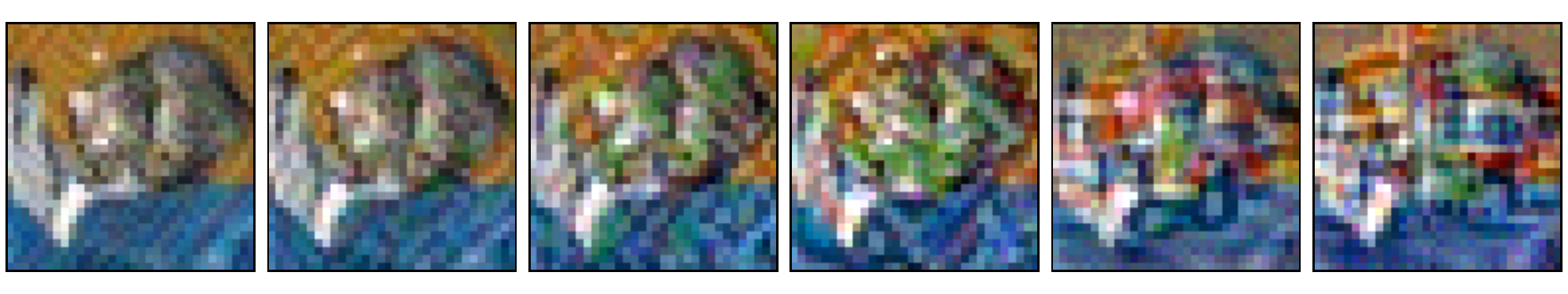}
	\end{center}
	\caption{Illustration of universal perturbations on CIFAR10 for sharedness $s=1$ (top row) and $s=64$ (middle row) for different values of $\varepsilon$. The bottom row shows a test image of a cat with the respective perturbation of the middle row being added.}
	\label{figure:cifar_perturbations}
\end{figure*}

Figure \ref{figure:cifar_perturbations} illustrates the minimal universal perturbation found for sharedness $s=1$ and $s=64$ for $\sigma=0.3$ and $\varepsilon \in \{2, 8, 14, 18, 26\}$. Universal perturbations of the undefended model resemble high-frequency noise and are quasi-imperceptible when added to an image. Shared adversarial training increases robustness and the resulting perturbations are more perceptible (for small $\varepsilon$) or even dominate the image: for larger $\varepsilon$, the cat in the figure is completely hidden and the perturbed image could also not be classified correctly by a human. Moreover, the perturbation becomes more structured and even object-like for larger $\varepsilon$. Note that the perturbations shown for $s=1$ also achieve high robustness but for smaller accuracy on clean data than those of shared adversarial training with $s=64$.

\subsection{Selection of subset of ImageNet} \label{section:subset_imagenet}
Since the generation of the Pareto fronts on the entire ImageNet dataset would be computationally very expensive, we restrict the experiment to a subset of ImageNet. We use classes defined in TinyImageNet to filter out the samples from ImageNet dataset. We conducted our experiments on the samples of 200 classes from ImageNet, which results in 258,601 train and 10,000 validation images.  Note that we take only the list of classes defined from TinyImageNet and use the data of those classes from ImageNet dataset with original resolution.

\subsection{Illustration of Universal Perturbations on ImageNet} \label{section:appendix_imagenet_perturbations}
\begin{figure*}[b]
	\begin{center}
		\includegraphics[width=\linewidth]{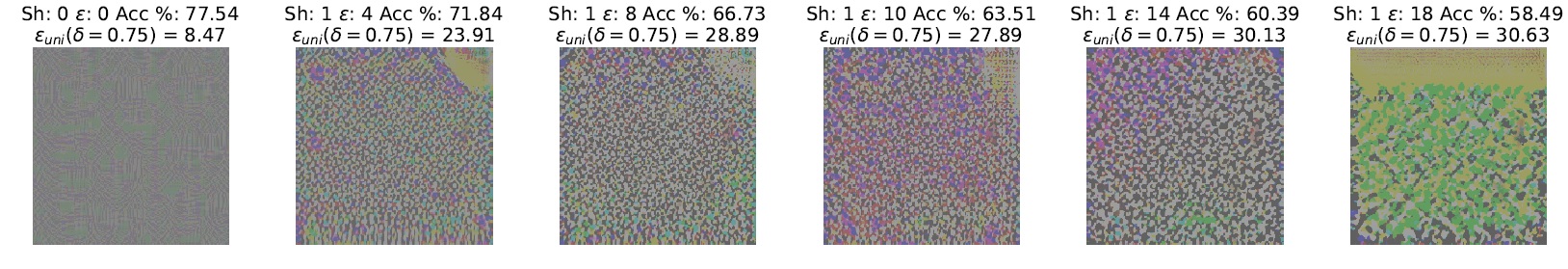}
		\includegraphics[width=\linewidth]{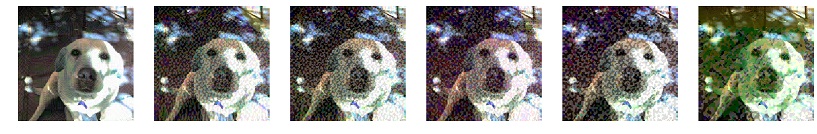}
		\includegraphics[width=\linewidth]{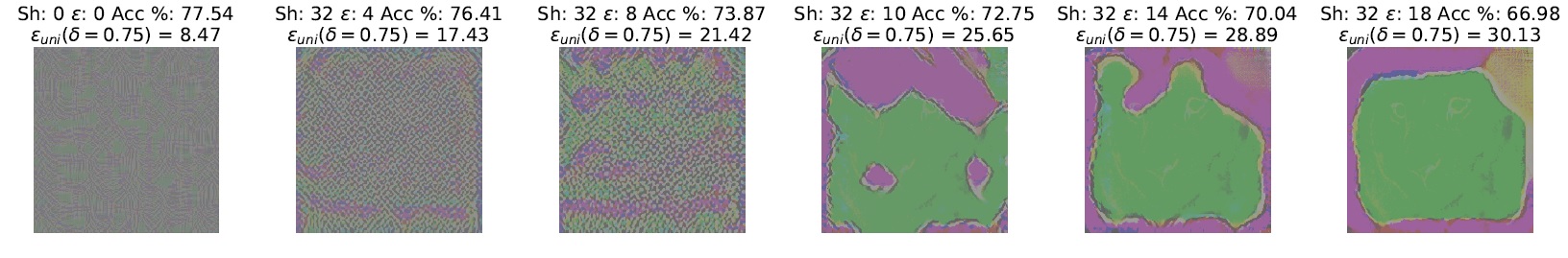}
		\includegraphics[width=\linewidth]{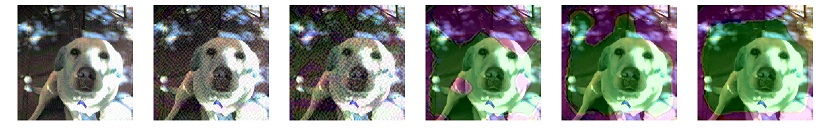}
	\end{center}
	\caption{Illustration of universal perturbations (not amplified) on ImageNet that are generated from different models with the settings: sharedness $s=1$ (top row) and $s=32$ (third row), $\sigma=1.0$ and different values of $\varepsilon$. The top-1 accuracy of the corresponding models and their smallest perturbation magnitude $\varepsilon$ that results in a misclassification rate of atleast $75\%$ are also shown. The second and bottom rows show a test image of a dog added with the respective universal perturbations from the first and third row. The models hardened with both standard and shared adversarial training demonstrate higher robustness when compared against the undefended model and universal perturbations become clearly visible. However, the shared adversarial training outperforms its counterpart in terms of robustness against perturbations and performance on unperturbed inputs. The perturbation of models from standard adversarial training resemble high frequency noise whereas the perturbations of the latter becomes much smoother for larger $\varepsilon$.}
	\label{figure:imagenet_perturbations}
\end{figure*}

We depict the universal perturbations with minimum magnitude on different models that are obtained from settings $\sigma=1.0$, sharedness $s \in \{1,32\}$ and different values of $\varepsilon$ on the subset of ImageNet in Figure \ref{figure:imagenet_perturbations}. It can be clearly seen that both the standard ($s=1$) and shared adversarial training ($s=32$) increase robustness when compared against the undefended model but the latter handles the trade-off between performance on unperturbed data and robustness more gracefully. The universal perturbations become clearly visible on a model hardened with shared adversarial training with only a marginal loss of  $5\%$ in top-1 accuracy and perturbations become much smoother for larger $\varepsilon$.

\subsection{Predicted Class after Untargeted Universal Perturbations} \label{section:appendix_imagenet_class_count}

\begin{figure*}[h]
	\begin{center}
		\includegraphics[width=\linewidth]{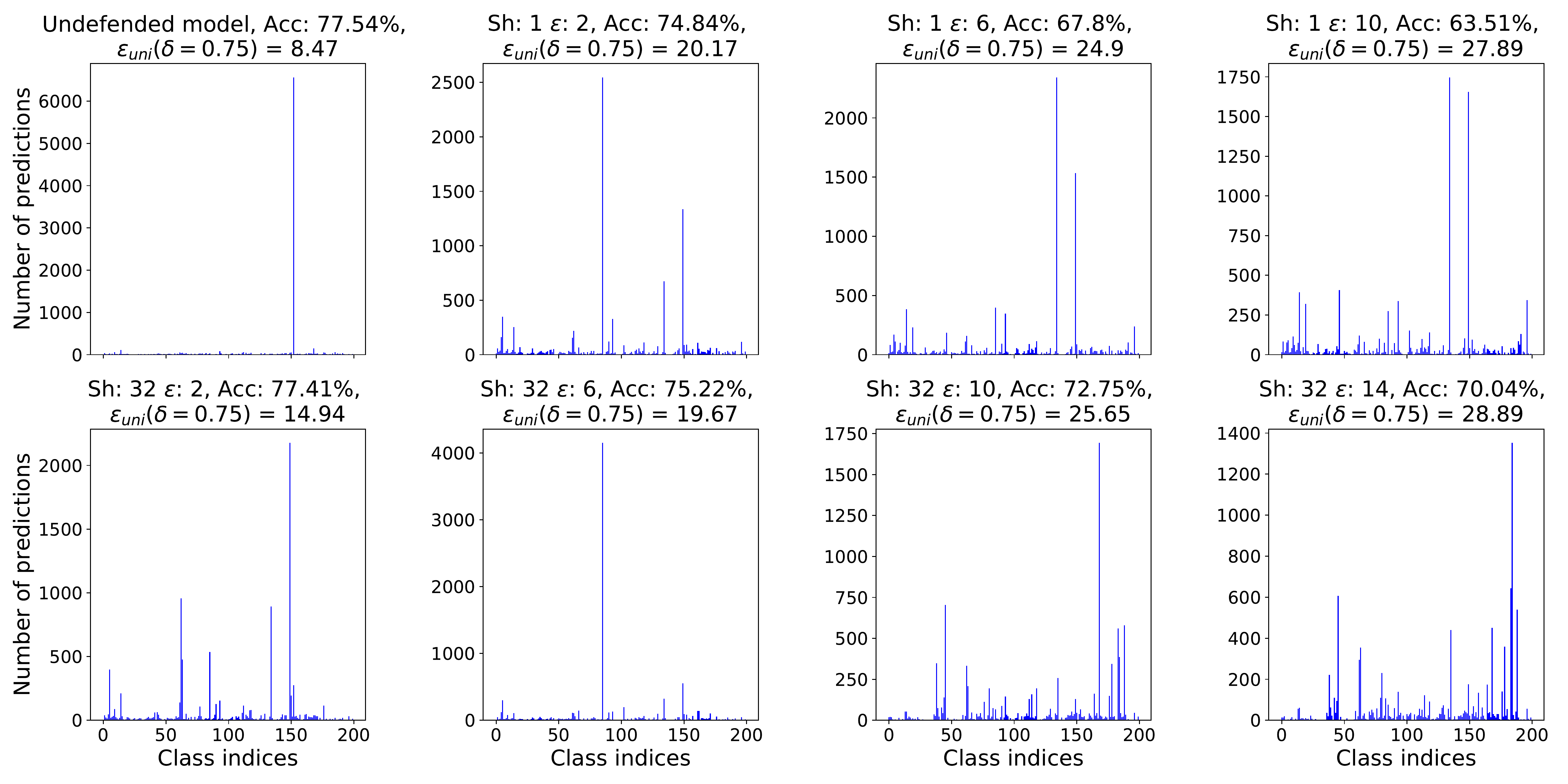}
	\end{center}
	\caption{The figure shows histogram of the predicted classes over validation data of different models when an untargeted universal perturbation is added. The histogram is based on 200 classes ImageNet validation data. In other words, a bar in each histogram represents the number of times a class (represented by class index) is predicted over the validation samples. It is interesting to note that the undefended model almost always misclassified the adversarial samples (samples added with universal perturbations) under the same class even though attack is untargeted. In contrast, the defended models from standard and shared adversarial training have higher entropy in their predictions.}
	\label{figure:class_count}
\end{figure*}

Figure \ref{figure:class_count} shows which class is predicted on ImageNet validation data after an untargeted universal perturbation (for the respective model) is added. While the undefended model predicts nearly always the same (wrong) class, the models defended with standard and shared adversarial training have a substantially higher entropy in their predictions. Prior work \cite{jetley_friends_2018} has also observed that undefended models typically misclassify images perturbed with universal perturbation to the same class even though the attack is untargeted. Based on this observation, they hypothesized that directions in which a classifier is vulnerable to universal perturbations coincide with directions important for correct prediction on unperturbed data. We believe it would be important to re-examine these results for a defended model.

\subsection{Attacks on Semantic Image Segmentation} \label{section:cityscapes_adv_examples}
\begin{figure*}[b!]
	\begin{center}
		\includegraphics[width=\textwidth]{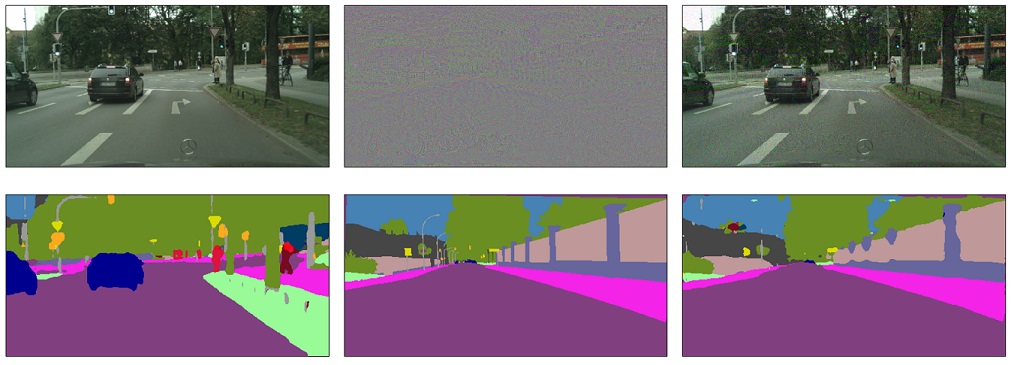}
	\end{center}
	\caption{Targeted universal perturbations on Cityscapes for a model pretrained with empirical risk minimization. The shown perturbation upper bounds the robustness of the model to $\varepsilon_{uni}(\delta=0.95) \leq 19.89$. Top row shows original image, universal perturbation, and perturbed image. Bottom row shows prediction on original image, target segmentation, and prediction on perturbed image.}
	\label{figure:cityscapes_up_targeted_erm}
\end{figure*}

\begin{figure*}[tb]
	\begin{center}
		\includegraphics[width=\textwidth]{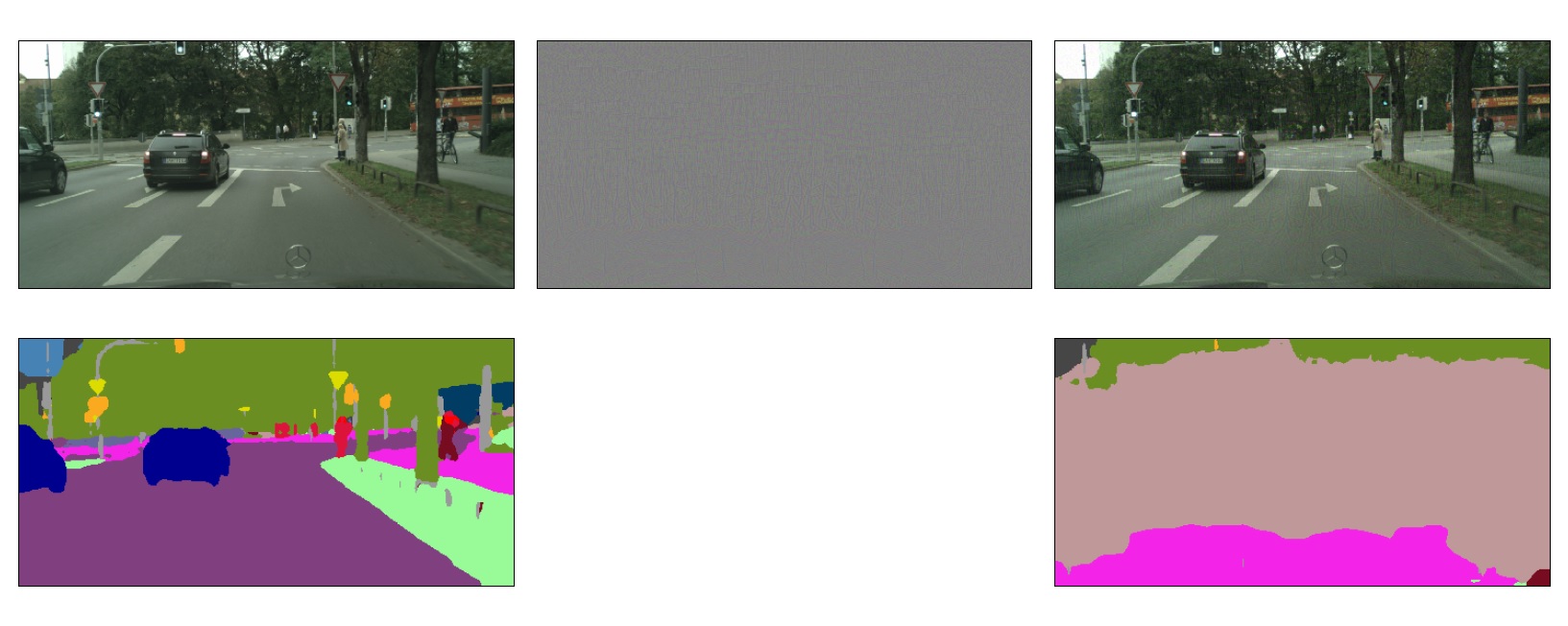}
	\end{center}
	\caption{Untargeted universal perturbations on Cityscapes for a model pretrained with empirical risk minimization. The shown perturbation upper bounds the robustness of the model to $\varepsilon_{uni}(\delta=0.95) \leq 8.5$. Top row shows original image, universal perturbation, and perturbed image. Bottom row shows prediction on original image and prediction on perturbed image.}
	\label{figure:cityscapes_up_untargeted_erm}
\end{figure*}

\begin{figure*}[tb]
	\begin{center}
		\includegraphics[width=\textwidth]{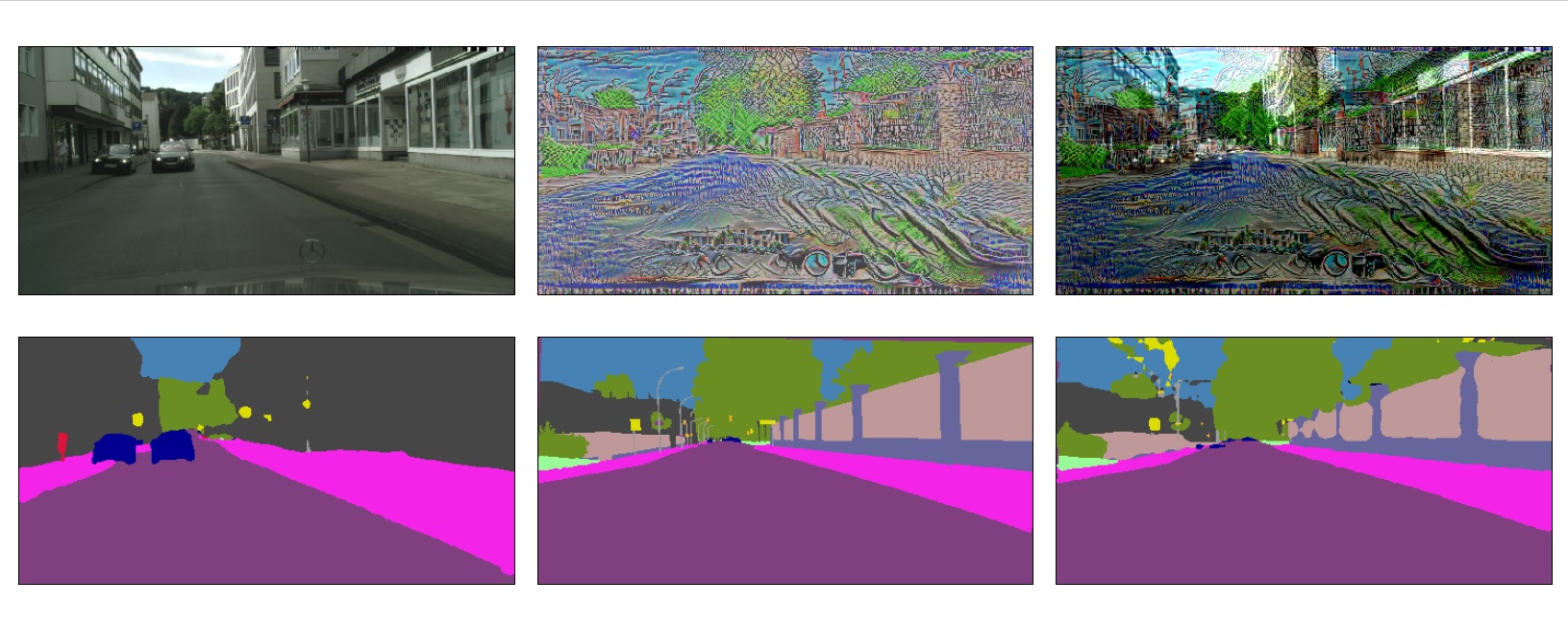}
	\end{center}
	\caption{Targeted universal perturbations on Cityscapes for a model trained with adversarial training. The shown perturbation upper bounds the robustness of the model to $\varepsilon_{uni}(\delta=0.95) \leq 62.5$. Top row shows original image, universal perturbation, and perturbed image. Bottom row shows prediction on original image, target segmentation, and prediction on perturbed image.}
	\label{figure:cityscapes_up_targeted_at}
\end{figure*}

\begin{figure*}[tb]
	\begin{center}
		\includegraphics[width=\linewidth]{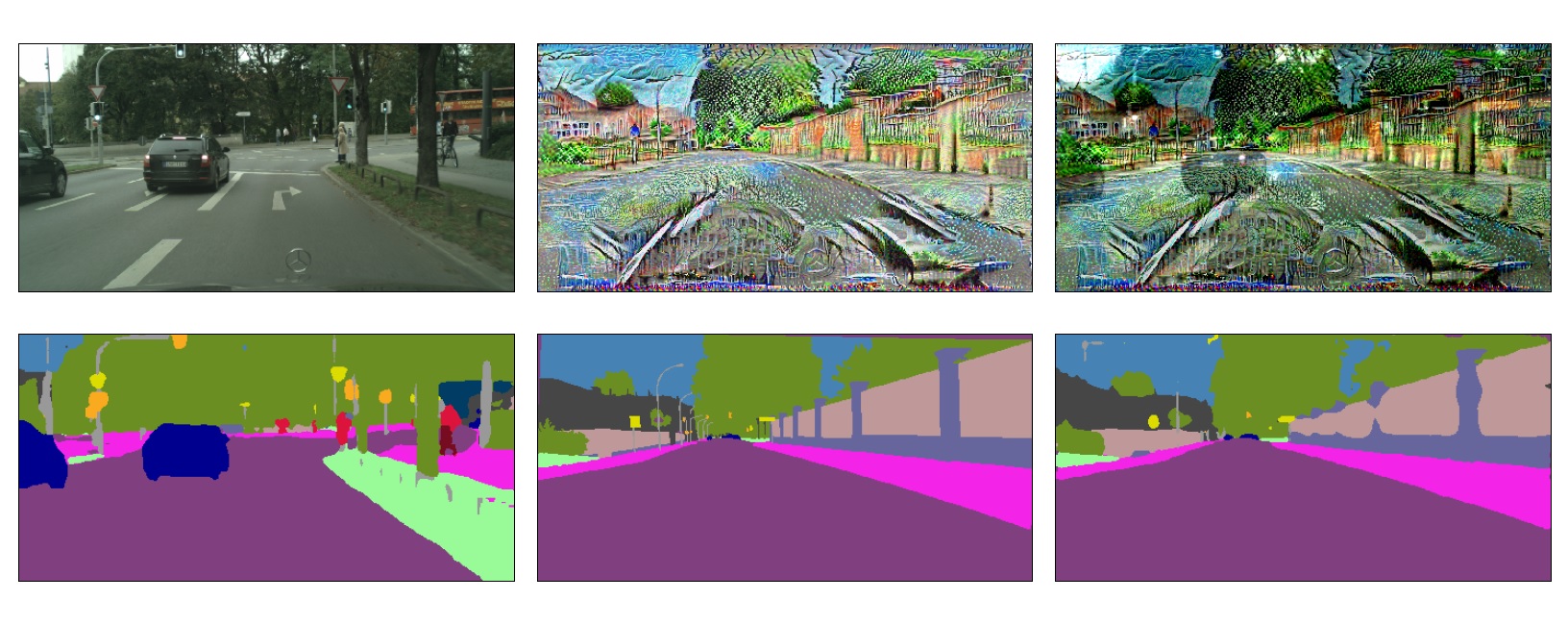}
	\end{center}
	\caption{Universal perturbations on Cityscapes for a model trained with shared adversarial training. The shown perturbation upper bounds the robustness of the model to $\varepsilon_{uni}(\delta=0.95) \leq 111.7$. Top row shows original image, universal perturbation, and perturbed image. Bottom row shows prediction on original image, target segmentation, and prediction on perturbed image.}
	\label{figure:cityscapes_up_targeted_sat}
\end{figure*}

\begin{figure*}[tb]
	\begin{center}
		\includegraphics[width=\textwidth]{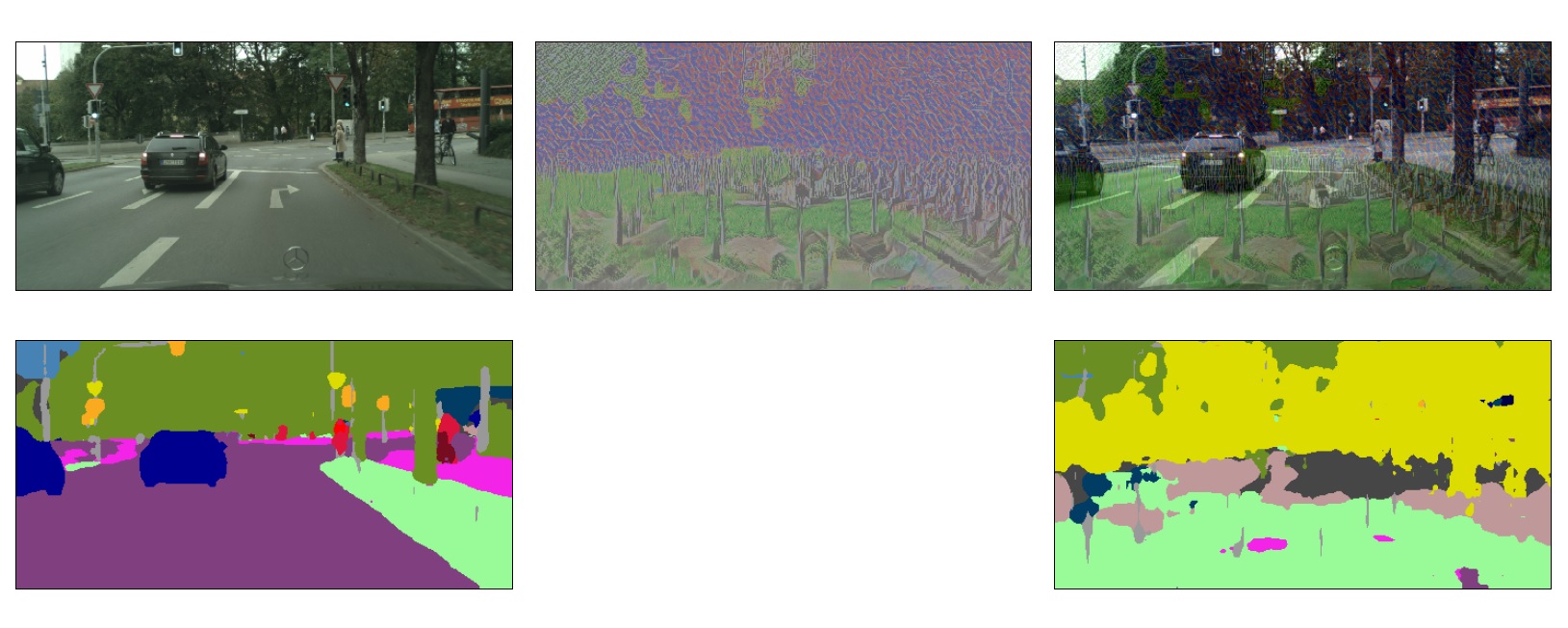}
	\end{center}
	\caption{Untargeted universal perturbations on Cityscapes for a model trained with adversarial training. The shown perturbation upper bounds the robustness of the model to $\varepsilon_{uni}(\delta=0.95) \leq 25$. Top row shows original image, universal perturbation, and perturbed image. Bottom row shows prediction on original image and prediction on perturbed image.}
	\label{figure:cityscapes_up_untargeted_at}
\end{figure*}

\begin{figure*}[tb]
	\begin{center}
		\includegraphics[width=\textwidth]{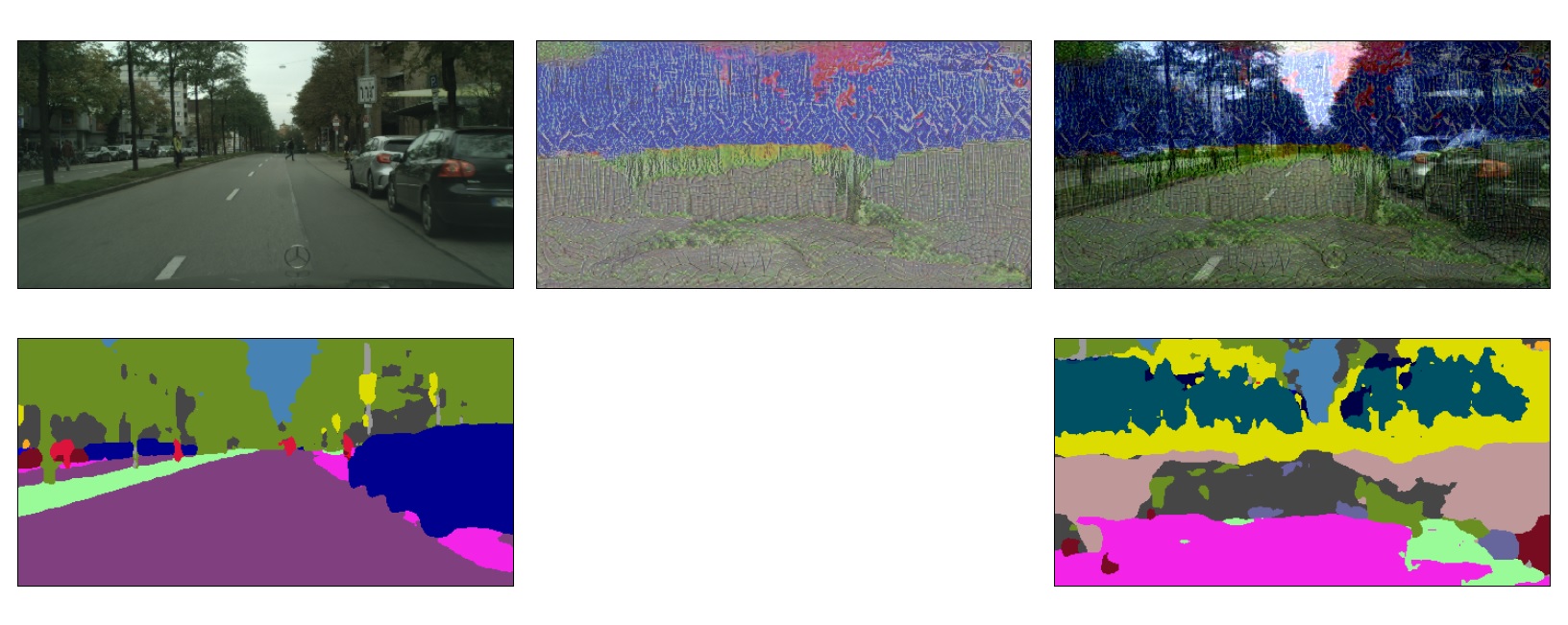}
	\end{center}
	\caption{Untargeted universal perturbations on Cityscapes for a model trained with shared adversarial training. The shown perturbation upper bounds the robustness of the model to $\varepsilon_{uni}(\delta=0.95) \leq 47.8$. Top row shows original image, universal perturbation, and perturbed image. Bottom row shows prediction on original image, target segmentation, and prediction on perturbed image.}
	\label{figure:cityscapes_up_untargeted_sat}
\end{figure*}

\begin{figure*}[tb]
	\begin{center}
		\includegraphics[width=.8\textwidth]{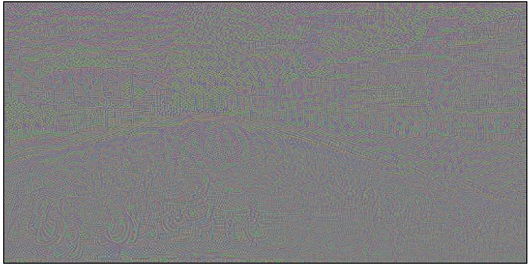}
		\includegraphics[width=.8\textwidth]{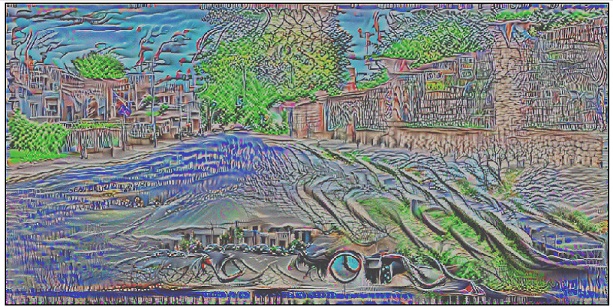}
		\includegraphics[width=.8\textwidth]{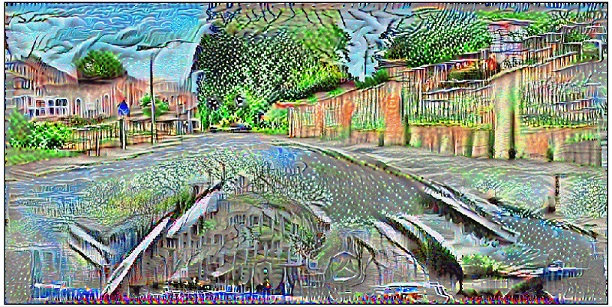}
	\end{center}
	\caption{Illustration of targeted universal perturbation for empirical risk minimization (top), adversarial training (middle), and shared adversarial training (bottom).}
	\label{figure:cityscapes_perturbations}
\end{figure*}

We illustrate universal perturbations for targeted and untargeted attacks on different models in this section. We illustrate the effect of the perturbations on one image; however, the perturbations are not specific for this image. 
For the model trained with empirical risk minimization, Figure \ref{figure:cityscapes_up_targeted_erm} shows a targeted attack and Figure \ref{figure:cityscapes_up_untargeted_erm} an untargeted attack. For the model trained with adversarial training, Figure \ref{figure:cityscapes_up_targeted_at} shows a targeted attack and Figure \ref{figure:cityscapes_up_untargeted_at} an untargeted attack. For the model trained with shared adversarial training, Figure \ref{figure:cityscapes_up_targeted_sat} shows a targeted attack and Figure \ref{figure:cityscapes_up_untargeted_sat} an untargeted attack. We also illustrate the universal perturbations found for different models on targeted attacks in Figure \ref{figure:cityscapes_perturbations}.

\begin{table*}
	\begin{center}
		\caption{Universal robustness and classification accuracy of ResNet20 trained with standard adversarial training ($s=1$) and \emph{shared adversarial training} $s \in \{8, 64\}$ against S-PGD universal perturbations on CIFAR10 under different range of attack parameters $\varepsilon$ and $\sigma$. The pictorial representation of these entries are depicted in Figure \ref{figure:cifar_pareto} (left). The bold entries represent the model trained with \emph{shared adversarial training} $s=64$ that yields threefold increase in robustness when compared to undefended model with a drop of less than $3.5\%$ in accuracy.} \label{tab:cifar10_robustness}
		\begin{tabular}{|c|c|c|M{11mm}|M{11mm}|M{11mm}|M{11mm}|M{11mm}|M{11mm}|M{11mm}|M{11mm}|M{11mm}|}
			
			$\sigma$ & $s$ && $\varepsilon=2$ &$\varepsilon=4$ & $\varepsilon=6$ & $\varepsilon=8$ &
			$\varepsilon=10$ &$\varepsilon=14$ & $\varepsilon=18$ & $\varepsilon=22$ & $\varepsilon=26$\\
			\hline
			\hline
			\multirow{ 2}{*}{0.3}
			& \multirow{2}{*}{1}
			& Acc.(\%)
			& 92.16  & 90.42  & 88.76  & 84.02  & 78.54  & 73.61  & 70.32  & 67.63  & 61.73  \\
			&& $\varepsilon_{uni}(\delta=0.75)$
			& 28.14  & 33.62  & 34.37  & 40.59  & 38.85  & 41.59  & 56.03  & 48.06  & 48.06  \\
			\cline{2-12}
			& \multirow{2}{*}{8}
			& Acc.(\%)
			& 93.04  & 92.05  & 91.34  & 90.75  & 89.09  & 85.87  & 82.68  & 79.69  & 76.35  \\
			&& $\varepsilon_{uni}(\delta=0.75)$
			& 23.41  & 26.89  & 32.62  & 34.12  & 32.87  & 40.59  & 45.82  & 47.31  & 54.54  \\
			\cline{2-12}
			& \multirow{2}{*}{64}
			& Acc.(\%)
			& 93.72  & 93.36  & 92.84  & 92.47  & 91.90  & 90.89  & 88.93  & 86.13  & 83.89  \\
			&& $\varepsilon_{uni}(\delta=0.75)$
			& 19.42  & 22.66  & 25.90  & 27.89  & 32.37  & 39.10  & 40.34  & 42.58  & 46.32  \\
			\hline 
			\hline
			
			\multirow{ 2}{*}{0.5}
			& \multirow{2}{*}{1}
			& Acc.(\%)
			& 91.77  & 89.70  & 87.60  & 85.44  & 82.64  & 71.81  & 66.61  & 64.46  & 62.16  \\
			&& $\varepsilon_{uni}(\delta=0.75)$
			& 30.63  & 38.10  & 39.35  & 42.33  & 47.31  & 61.26  & 55.78  & 93.27  & 63.75  \\
			\cline{2-12}
			& \multirow{2}{*}{8}
			& Acc.(\%)
			& 92.62  & 91.69  & 90.86  & 90.11  & 89.04  & 86.16  & 82.92  & 80.54  & 75.99  \\
			&& $\varepsilon_{uni}(\delta=0.75)$
			& 23.16  & 29.88  & 33.87  & 36.61  & 38.10  & 42.83  & 43.83  & 53.54  & 64.00  \\
			\cline{2-12}
			& \multirow{2}{*}{64}
			& Acc.(\%)
			& 93.55  & 93.09  & 92.50  & 91.99  & 91.62  & 90.58  & 88.52  & 86.65  & 84.13  \\
			&& $\varepsilon_{uni}(\delta=0.75)$
			& 20.67  & 23.91  & 28.14  & 29.88  & 38.10  & 37.60  & 41.34  & 47.31  & 51.05  \\
			\hline 
			\hline
			
			\multirow{ 2}{*}{0.7}
			& \multirow{2}{*}{1}
			& Acc.(\%)
			& 91.55  & 89.07  & 86.58  & 84.47  & 81.97  & 78.75  & 73.29  & 64.15  & 63.63  \\
			&& $\varepsilon_{uni}(\delta=0.75)$
			& 31.63  & 42.08  & 39.84  & 45.82  & 46.82  & 54.04  & 63.75  & 95.63  & 89.65  \\
			\cline{2-12}
			& \multirow{2}{*}{8}
			& Acc.(\%)
			& 92.51  & 91.29  & 90.23  & 89.12  & 88.08  & 85.70  & 83.74  & 80.27  & 78.00  \\
			&& $\varepsilon_{uni}(\delta=0.75)$
			& 26.40  & 30.38  & 35.11  & 39.84  & 44.08  & 45.07  & 45.82  & 50.55  & 54.54  \\
			\cline{2-12}
			& \multirow{2}{*}{64}
			& Acc.(\%)
			& 93.34  & 92.81  & 92.27  & 91.67  & 91.30  & \textbf{89.94}  & 88.25  & 85.96  & 83.94  \\
			&& $\varepsilon_{uni}(\delta=0.75)$
			& 20.67  & 23.91  & 27.89  & 30.88  & 36.11  & \textbf{44.08}  & 43.58  & 51.05  & 50.80  \\
			\hline 
			\hline
			
			\multirow{ 2}{*}{0.9}
			& \multirow{2}{*}{1}
			& Acc.(\%)
			& 91.45  & 88.54  & 85.51  & 82.97  & 80.38  & 76.35  & 72.94  & 70.14  & 68.20  \\
			&& $\varepsilon_{uni}(\delta=0.75)$
			& 32.87  & 38.85  & 46.07  & 50.55  & 54.54  & 58.27  & 63.25  & 64.50  & 59.77  \\
			\cline{2-12}
			& \multirow{2}{*}{8}
			& Acc.(\%)
			& 92.27  & 90.97  & 89.89  & 88.63  & 86.89  & 84.25  & 81.63  & 79.13  & 76.56  \\
			&& $\varepsilon_{uni}(\delta=0.75)$
			& 26.40  & 33.12  & 37.10  & 40.34  & 44.08  & 45.82  & 47.81  & 55.28  & 58.02  \\
			\cline{2-12}
			& \multirow{2}{*}{64}
			& Acc.(\%)
			& 93.18  & 92.61  & 92.12  & 91.42  & 90.85  & 89.32  & 87.41  & 85.26  & 83.10  \\
			&& $\varepsilon_{uni}(\delta=0.75)$
			& 22.16  & 25.40  & 29.63  & 32.87  & 36.61  & 43.33  & 46.07  & 45.32  & 54.04  \\
			\hline 
		\end{tabular}
	\end{center}
\end{table*}

\begin{table*}
	\begin{center}
		\caption{Universal robustness and classification accuracy of WRN-50-2-bottleneck trained with standard adversarial training ($s=1$) and \emph{shared adversarial training} ($s=32$) against S-PGD universal perturbations on a subset of ImageNet (200 classes) under different range of attack parameters $\varepsilon$ and $\sigma$. The pictorial representation of these entries are depicted in Figure \ref{figure:tinyimagenet_pareto}. The bold entries represent the model trained with \emph{shared adversarial training} $s=32$ that yields threefold increase in robustness when compared to undefended model with a drop of less than $5\%$ in accuracy.}\label{tab:imagenet_robustness}
		\begin{tabular}{|c|c|c|M{11mm}|M{11mm}|M{11mm}|M{11mm}|M{11mm}|M{11mm}|M{11mm}|M{11mm}|M{11mm}|}
			
			$\sigma$ & $s$ && $\varepsilon=2$ &$\varepsilon=4$ & $\varepsilon=6$ & $\varepsilon=8$ &
			$\varepsilon=10$ &$\varepsilon=14$ & $\varepsilon=18$ & $\varepsilon=22$ & $\varepsilon=26$\\
			\hline
			\hline
			\multirow{ 2}{*}{0.5}
			& \multirow{2}{*}{1}
			& Acc.(\%)
			& 75.81 & 74.49 & 71.14 & 71.48 & 68.78 & 66.25 & 64.05 & 59.04 & 57.54 \\
			&& $\varepsilon_{uni}(\delta=0.75)$
			& 17.92 & 18.92 & 23.90 & 17.92 & 13.94 & 20.66 & 20.41 & 29.38 & 25.89 \\
			\cline{2-12}
			& \multirow{2}{*}{32}
			& Acc.(\%)
			& 77.50 & 77.23 & 75.46 & 72.77 & 68.43 & 64.21 & 56.48 & 54.58 & 48.30 \\
			&& $\varepsilon_{uni}(\delta=0.75)$
			& 13.44 & 15.93 & 17.43 & 12.70 & 18.67 & 28.13 & 33.36 & 43.33 & 45.57 \\
			\hline
			\hline
			
			\multirow{2}{*}{1.0}
			& \multirow{2}{*}{1}
			& Acc.(\%)
			& 74.83 & 71.84 & 67.79 & 66.72 & 63.51 & 60.38 & 58.49 & 57.23 & 55.19 \\
			&& $\varepsilon_{uni}(\delta=0.75)$
			& 20.17 & 23.90 & 24.90 & 28.88 & 27.89 & 30.13 & 30.62 & 32.12 & 32.37 \\
			\cline{2-12}
			& \multirow{2}{*}{32}
			& Acc.(\%)
			& 77.41 & 76.41 & 75.22 & 73.86 & \textbf{72.74} & 70.04 & 66.97 & 62.36 & 56.99 \\
			&& $\varepsilon_{uni}(\delta=0.75)$
			& 14.94 & 17.43 & 19.67 & 21.41 & \textbf{25.64} & 28.88 & 30.13 & 29.38 & 33.36 \\
			\hline
			
		\end{tabular}
	\end{center}
\end{table*}

\end{document}